\documentclass[10pt,journal,compsoc]{IEEEtran}
\usepackage{lineno}
\usepackage{epsfig}
\usepackage{chngpage}
\usepackage{float}
\usepackage{amsmath}
\usepackage{graphicx}
\usepackage{array}
\usepackage{diagbox}
\usepackage{epstopdf}
\usepackage{algorithm}
\usepackage{algorithmicx}
\usepackage{algpseudocode}
\usepackage{threeparttable}
\usepackage{subfigure}
\usepackage{multirow}
\usepackage{longtable}
\usepackage{bm}
\usepackage[table,xcdraw]{xcolor}
\usepackage{url}
\usepackage{booktabs}
\usepackage{multirow}
\usepackage[colorlinks,linkcolor=blue]{hyperref}
\usepackage{algorithm}
\usepackage{algorithmicx}
\usepackage{algpseudocode}
\usepackage{amsmath}
\usepackage{amssymb}
\newtheorem{myDef}{Definition}

\begin{document}

\title{Shifu2: A Network Representation Learning Based Model for Advisor-advisee Relationship Mining}
\author{Jiaying Liu,
        Feng Xia,~\IEEEmembership{Senior Member,~IEEE,}
        Lei Wang,
        Bo Xu,
        Xiangjie Kong,~\IEEEmembership{Senior Member,~IEEE,}
        Hanghang Tong,
        and Irwin King,~\IEEEmembership{Fellow,~IEEE}

\IEEEcompsocitemizethanks{\IEEEcompsocthanksitem J. Liu, L. Wang, B. Xu, and X. Kong are with the Key Laboratory for Ubiquitous Network and Service Software of Liaoning Province, School of Software, Dalian University of Technology, China.

\IEEEcompsocthanksitem F. Xia is with School of Science, Engineering and Information Technology, Federation University Australia, Australia, and School of Software, Dalian University of Technology, China.

\IEEEcompsocthanksitem H. Tong is with Department of Computer Science, University of Illinois at Urbana-Champaign, USA.

\IEEEcompsocthanksitem I. King is with Department of Computer Science and Engineering, The Chinese University of Hong Kong, Hong Kong.}
\thanks{Corresponding author: Feng Xia; email: f.xia@ieee.org.}}

\markboth{IEEE Transactions on Knowledge and Data Engineering,~Vol.~0, No.~0, October~2019}%
{Shell \MakeLowercase{\textit{et al.}}: Bare Demo of IEEEtran.cls for Computer Society Journals}
%



\IEEEtitleabstractindextext{%
\begin{abstract}
  The advisor-advisee relationship represents direct knowledge heritage, and such relationship may not be readily available from academic libraries and search engines. This work aims to discover advisor-advisee relationships hidden behind scientific collaboration networks. For this purpose, we propose a novel model based on Network Representation Learning (NRL), namely Shifu2, which takes the collaboration network as input and the identified advisor-advisee relationship as output. In contrast to existing NRL models, Shifu2 considers not only the network structure but also the semantic information of nodes and edges. Shifu2 encodes nodes and edges into low-dimensional vectors respectively, both of which are then utilized to identify advisor-advisee relationships. Experimental results illustrate improved stability and effectiveness of the proposed model over state-of-the-art methods. In addition, we generate a large-scale academic genealogy dataset by taking advantage of Shifu2.
\end{abstract}

}

\maketitle

\IEEEdisplaynontitleabstractindextext

\IEEEpeerreviewmaketitle

\IEEEraisesectionheading{\section{Introduction}\label{sec1}}
\IEEEPARstart{I}{t} is well-known that academic network can be formed according different types of relationships, such as colleagues, friends, and advisor-advisee relationships. These relationships usually reflect different interpersonal interactions. For example, in advisor-advisee relationships, a PhD candidate's research topic is usually determined by his/her advisor (i.e., supervisor). While in friendships, one's daily schedule may be reflected by his/her friends. These interactions govern the dynamics and the complexity of social networks. To better model the interaction based on Network Science, a concrete network is abstracted into a graph consisting of nodes and edges, where nodes represent the entities and the edges indicate the different relationships. Hence, we can model the influence of nodes and edges from both local and global perspectives using graph theoretical methods and machine learning techniques.

With the rapid growth of scholarly information and artificial intelligence~\cite{liu2018artificial}, researchers have shown rapidly-increasing interest in exploring big scholarly data (BSD)~\cite{xia2017big}. BSD contains not only scholarly records including papers, authors, venues, and citations, but also other related data such as scientific networks and digital libraries. Obviously, the data can form various types of networks within the science of science~\cite{fortunato2018science}. For example, in an academic collaboration network, nodes usually represent scholars and edges mean that connected scholars have even collaborated with each other. Another important type, the citation network, contains a large number of papers (represented as nodes) and citation relationships (represented as edges). Based on the analysis of these networks, we can effectively retrieve useful information hidden behind the nodes. Newman et al.~\cite{newman2001structure} analyze the scientific collaboration networks and highlight the differences in collaboration patterns between different research fields. Wang et al.~\cite{wang2017scientific} predict scientific collaboration sustainability from the perspectives of collaboration times and collaboration duration. Yu et al.~\cite{yu2017team} recognize academic collaborative teams by exploring collaboration intensity.

As a relationship in scientific collaboration networks, the advisor-advisee relationship is vitally important for scholars. Generally speaking, each newcomer in academia will be advised/supervised by an advisor. The hypothesis that both advisees and advisors benefit from advisor-advisee relationships is proved by copious literature~\cite{chao1992formal,liu2018understanding}. Specifically, from the advisees' perspective, they receive guidance and support (i.e., funds for scientific research, career advice). Malmgren et al.~\cite{malmgren2010role} also point out that the advisor-advisee relationship is important to academic organizations because advisees usually would like to be committed to their organizations after graduation. Therefore, identifying advisor-advisee relationships will benefit many significant applications, such as double-blind peer review, scientific genealogy generation, and scientific career analysis. More importantly, mining and analyzing of advisor-advisee relationships will help us better understand underlying principles of the academic society from the perspectives of collaborative teamwork formation~\cite{beaver2001reflections,lariviere2015team} and scientific entities modeling~\cite{pan2018memory,liu2018hot,he2018modeling}.

In practice, the lack of high-quality data about academic mentorships significantly limits the exploration of advisor-advisee relationships. There are some projects that aim at helping users track the academic genealogy, such as the Mathematics Genealogy Project\footnote{https://www.genealogy.math.ndsu.nodak.edu/}, Neurotree\footnote{https://neurotree.org/neurotree/}, and the MPACT Project\footnote{http://www.ibiblio.org/mpact/}. But all of projects rely on manual processing, which limits the number of records. Meanwhile, few of them provide scientific related information for advisor-advisee pairs, such as publication records and collaboration information. Although some researchers~\cite{wang2017shifu,wang2010mining,zhao2018identifying} have proposed the methods to deduce mentorships from academic publication networks, they ignore dynamics and complexity in science itself. It is still an open issue to develop appropriate methods that can extract advisor-advisee relationships automatically from scientific digital libraries. However, it suffers from a number of challenges due to the inherent properties and environmental attributes of the relationship. For instance, the relationship is time-dependent. The identity of an advisee will change over time. How to control this time-dependent factor is critical to the solution.

\begin{figure}
  \centering
  \includegraphics[width=0.5\textwidth]{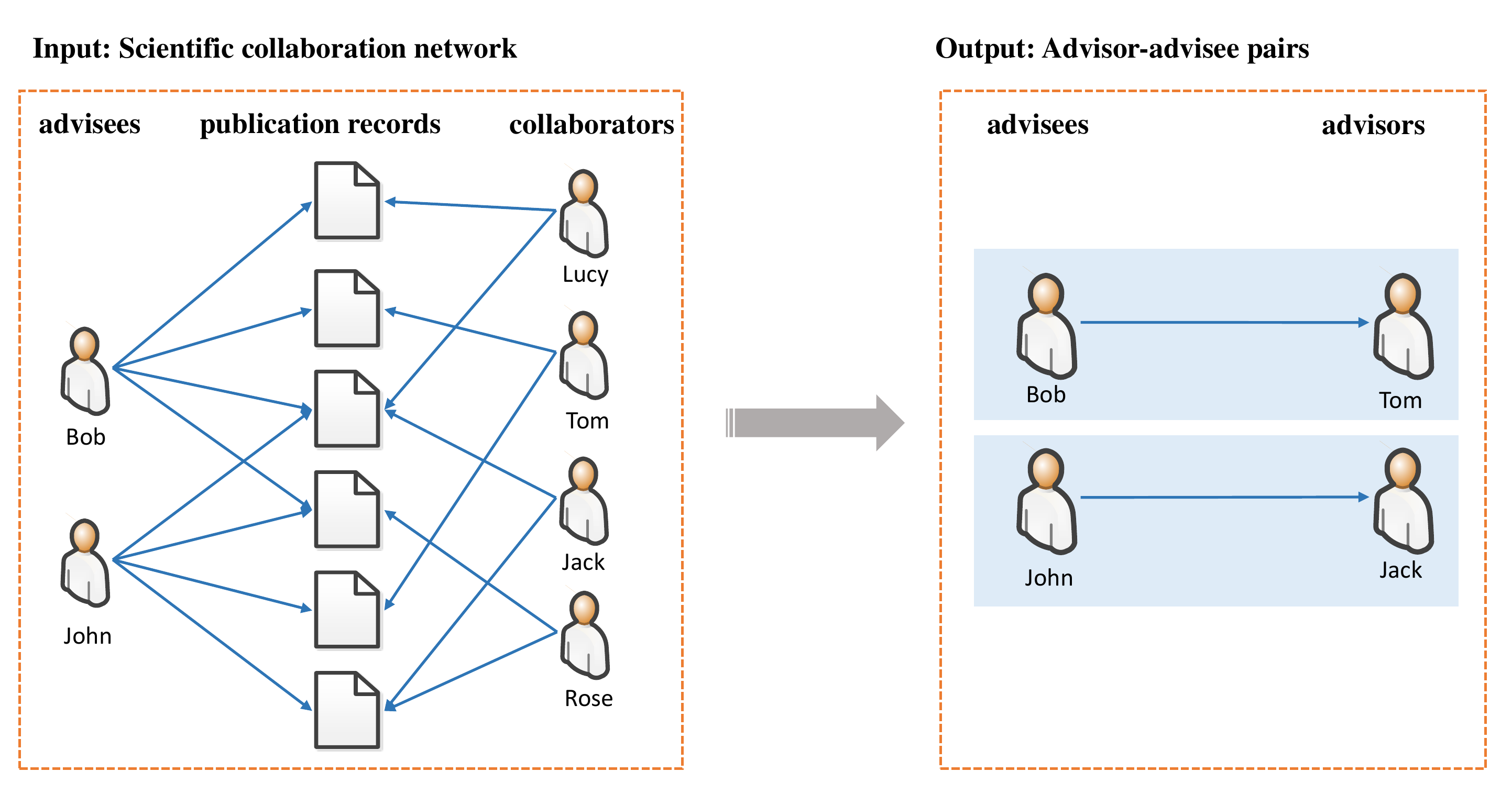}\\
  \caption{An example of advisee-advisor relationship mining in the scientific collaboration network.}
  \label{fig:1}
\end{figure}
To tackle the above challenges, in this paper, we propose an effective model named Shifu2, which takes advantage of Network Representation Learning (NRL) techniques to form an intelligent solution for mining advisor-advisee relationships hidden in a scientific collaboration network. To better elaborate the problem, Fig.~\ref{fig:1} presents an example of the advisor-advisee relationship identification in the network. Shifu2 is an NRL model. Specifically, we regard the advisor-advisee relationship mining problem as a classification problem. The collaborators and the attributes of each scholar are modeled as joint vectors. We design an efficient algorithm to optimize the process of vector representation composed of edge attributes and node attributes. We conduct extensive experiments to evaluate the effectiveness of our model. The results illustrate that the best performance of this method can achieve the accuracy of 92\%, leading by at least 5\% against the baseline methods.

Furthermore, we analyze the differences between advisor-advisee relationships over varying intervals of time. In the application part, we consider these differences and add an age adjustment parameter and a disciplinary adjustment parameter. Thus, it can be applied to the entire dataset which contains multi-disciplinary publication records through a long history.

In summary, the main contributions of this paper can be summarized as follows:
\begin{itemize}
  \item \textbf{A Novel Mining Model}. We devise Shifu2 based on the NRL technique, to identify advisor-advisee relationships hidden in the scientific collaboration network. Unlike existing research, we consider semantic information of both nodes and edges for embedding. Experimental results demonstrate outstanding capabilities of Shifu2 for the task.
  \item \textbf{New Knowledge.} We discover the differences between each discipline with respect to the structures of advisor-advisee collaboration networks. Based on the obtained observations, we add the parameters of time adjustment and disciplinary adjustment to make Shifu2 universal. Consequently it can eliminate temporal/chronological differences as well as disciplinary differences.
  \item \textbf{A Benchmark Dataset}. By applying Shifu2 onto the entire pre-processed dataset (i.e., Microsoft Academic Graph (MAG)\footnote{https://www.openacademic.ai/oag/}), we generate a large-scale dataset containing not only advisor-advisee pairs but also the academic attributes and publication records of each scholar.
\end{itemize}

The rest of the paper is organized as follows. Section~\ref{sec2} lays out the research scope and introduces related work. Section~\ref{sec3} formally formulates the problem and presents the overall architecture of the proposed solution. Section~\ref{sec4} describes experimental settings and presents the results to illustrate the effectiveness of Shifu2. Besides, we also analyze the results and present the findings focusing on the application and visualization. Finally, Section~\ref{sec5} concludes the paper.

\section{Related Work}
\label{sec2}
The related work is divided into three parts. The first part deals with relation extraction (RE) in social networks. The second part reviews existing work for the advisor-advisee relationship identification. The third part focuses on NRL techniques.
\subsection{Social Relation Extraction}
RE is an important topic in knowledge graphs, which focuses on extracting relationships among entities to enrich existing information. In modeling relationships in knowledge graphs, relationships between entities are usually described as ``head" + ``relation" = ``tail", i.e., ``advisors" + ``advising" = ``advisees". Social relationship extraction (SRE) is an important subtask of RE, which focuses on mining social interrelationships between entities. Generally, the social network is defined as $G = (V, E)$, where $V$ means objectives and $E \subseteq (V \times V)$ are edges between them. The edges $E$ consists of two types, labeled $E_{L}$ and unlabeled $E_{U}$. Then the problem of SRE can be described as: How to predict the labels over each edge in $E_{U}$ based on $G$ and $E_L$?

The most commonly used techniques in SRE can be classified into four categories including similarity measures, statistical relational learning measures, graph mining measures, and machine learning techniques. Previous studies in SRE mainly use these techniques for unstructured data (e.g., text data, web pages, and the corpus of literature). Up to now, a number of studies have been carried out to identify intimate relationships. Diehl et al.~\cite{diehl2007relationship} utilize a supervised ranking approach to identify the manager-subordinate relationship. Reviewing the development of SRE techniques, the major problems are:

(1) Relationships in real-world networks are complex and dynamic, which means that entities will play different roles under different scenarios. The interactions between entities also change over time. A single label can not describe vertices well because it can provide neither sufficient descriptions nor dynamic characteristics.

(2) In comparison with RE in knowledge graphs where relations are well pre-defined with human efforts, relations between vertices in social networks are usually invisible.
We should consider how to obtain invisible information with the help of existing techniques.

\subsection{Advisor-advisee Relationship Identification}
The advisor-advisee relationship is one of the most important relationships in scientific collaboration networks. It can benefit many academic related applications such as advisor recommendation and reviewer recommendation. However, identification of such relationship is not straight forwards and the study still remains a challenge. Wang et al.~\cite{wang2010mining} propose TPFG to extract the advisor-advisee relationship based on the probabilistic factor graph. Zhao et al.~\cite{zhao2018identifying} capitalize on a deep model equipped with improved Refresh Gate Recurrent Units to identify the relationship. Another framework to identify the advisor-advisee relationship proposed by Wang et al.~\cite{wang2017shifu} exploits a deep learning method.

Shifu2 is built on top of a previous work named Shifu~\cite{wang2017shifu}. In the process of determining edge and node attributes, we refer to the feature selection part of Shifu. However, compared with Shifu, Shifu2 has further improvement in model selection, data processing, and application perspectives. Specifically, the main differences between Shifu2 and Shifu are:

(1) Shifu utilizes the Digital Bibliography \& Library Project (DBLP)\footnote{https://dblp.uni-trier.de/} dataset, which only contains publication records in the field of Computer Science. It means that the task is limited to the field of Computer Science. In Shifu2, we use the MAG dataset containing paper information for more fields. We crawl the real advisor-advisee pairs in six major fields including Chemistry, Computer Science, Economics, Engineering, Mathematics, and Physics. Through individual training of data in different fields, Shifu2 model has better applicability and scalability.

(2) Shifu utilizes supervised learning using a deep learning based model. Compared with Shifu, Shifu2 is an NRL model and represents each node in a low-dimensional space first. Then it capitalizes on the representations as the input of the task. As a result, Shifu2 is able to apply in large-scale networks and archive good performance in a short time. Shifu2 can better promote the convergence and reasoning, especially when labeled data is sparse. In addition, Shifu2 integrates attributes into node and edge attributes, which can better verify the effect of different attributes on the model.

(3) In the application part, the authors directly apply Shifu model on the large-scale network. On the contrary, Shifu2 preprocesses author name disambiguation, disciplinary differences elimination, and re-scaled number of publications to solve the problems of author name duplication, disciplinary differences, and temporal effect, respectively.

\subsection{Network Representation Learning}
With the development of machine learning techniques, how to represent nodes information in various networks has become a critical issue. NRL is becoming a key instrument in modeling network structure. It learns the representations for nodes and adopts them as features for downstream applications such as link prediction and node classification. Formally, NRL
learns a real vector $R_{v} \in R^{k}$ for each node, where the dimension of $k$ is much smaller than the total number of nodes $|V|$. The learning process can either be unsupervised or semi-supervised.

Existing NRL models can be broadly divided into two categories: one is based on network structures and the other considers external information such as text information and label categories. DeepWalk~\cite{perozzi2014deepwalk} and LINE~\cite{tang2015line} are typical models which attempt to learn the representations based on local network structures. To preserve the global network structure, Cao et al.~\cite{cao2015grarep} present GraRep to learn representations in weighted graphs. Wang et al.~\cite{wang2017community} propose M-NMF to preserve both the microscopic and community structure.

Recently, research on incorporating heterogeneous information into NRL models has received increasing interests~\cite{yang2015network,kim2017distinguishing,tu2017cane,tu2017transnet}. Among them, specific text information (e.g., labelling information) is usually used as edges' external information for node representation. Yang et al.~\cite{yang2015network} introduce TADW, which incorporates text information into
NRL models based on matrix factorization. CANE~\cite{tu2017cane} learns the context-aware embeddings for vertices based on the attention mechanism to model the semantic relations between themselves.

\section{Design of Shifu2}
\label{sec3}
\begin{figure*}
  \centering
  \includegraphics[width=\textwidth]{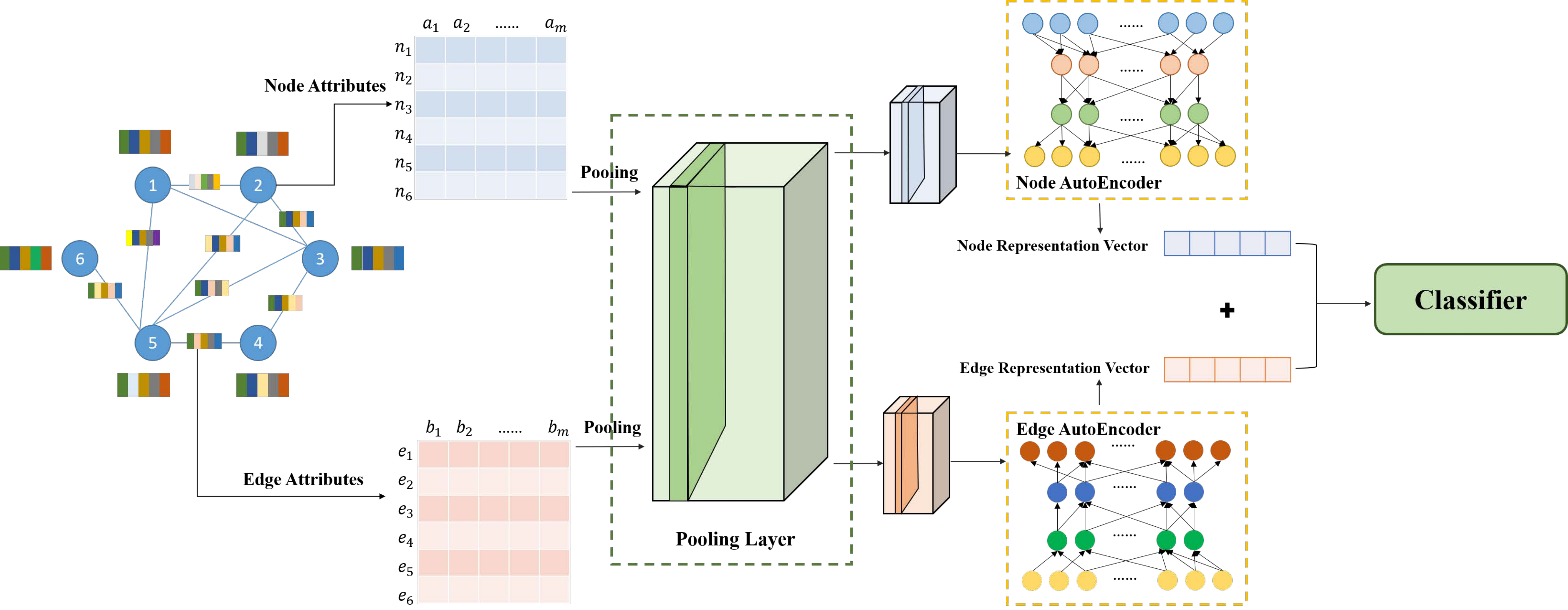}\\
  \caption{Framework of Shifu2, which contains four key components: (1) edge representation construction, (2) node representation construction, (3) pooling, and (4) classification. }
  \label{fig:2}
\end{figure*}
In this paper, we focus on mining advisor-advisee relationships hidden in the scientific collaboration network. Although NRL techniques can effectively learn the complex structures of networks, it is easy to make the representations sub-optimal because shallow network embedding models cannot capture the highly nonlinear network structure. Taking advantage of deep learning models which have great ability to learn the complex network structure~\cite{bengio2009learning,krizhevsky2012imagenet}, we adopt deep learning models to learn the representations for the target network.

Unlike existing studies, we incorporate attributes for both edges and nodes into NRL models and formalize the task as modeling the relations between nodes. In an academic collaboration network, both nodes and edges attributes can reflect network properties. For example, from the perspective of node attributes (i.e., academic age), advisors often have a longer career than their advisees. At the same time, edge attributes such as collaboration times can reflect collaboration intensity. In order to preserve network structures in the joint space composed of node attributes and edge attributes, we exploit the proximity jointly by using a deep learning model. As shown in Fig.~\ref{fig:2}, Shifu2 contains four critical components, i.e., edge representation construction part, node representation construction part, pooling part, and classification part. The node autoencoder is used to capture the local network structure and the edge autoencoder is used to preserve the collaboration information.

To clearly illustrate the problem, we first define the notations used in Shifu2 framework. Then we present the details of how to construct the representations for nodes and edges, respectively. At last, we introduce the overall reconstruction mechanism of Shifu2.

\subsection{Problem Formulation}
Since we focus on the collaboration in advisor-advisee relationships, we select all papers published by target scholars to construct the collaboration network ${G} = (V, E)$, where the node $v_i \in V$ represents the scholar $i$ and edges in $E$ weighted by collaboration times represent co-author relationships. Here we suppose that the advisor of $i$ is one of $i$'s collaborators, and we use $v_j\in C_i$ to represent his/her collaborator, where $C_i$ is the set of $i$'s collaborators. Shifu2 considers external information of edges and nodes at the same time.

\begin{myDef}
\textbf{Node Attributes}. Node attributes represent the inherent attributes of nodes. In the proposed model, it can be regarded as the local properties of scholars, such as the institution, the academic age, the number of publications, etc. We use node attributes to reflect the scholars' academic performance and the similarity of two scholars. For example, an advisor and his/her advisee are more likely to belong to the same institution, while the advisor may publish more papers than his/her students. \bm{$A_{n}$} $\in \mathbb{R}^{{{n_1} \times m_{1}}}$ preserves the node attributes, where $m_{1}$ is node attribute categories and $n_{1} = |V|$ is the total number of nodes in the network.
\end{myDef}

\begin{myDef}
\textbf{Edge Attributes}. Edge attributes aim to depict relationship intensity or other properties. In Shifu2, they are used to describe the collaboration intensity between scholars, such as collaboration times, collaboration duration and so on. In contrast to  other collaborators, the collaboration pattern between advisees and advisors are special. For example, advisees tend to collaborate frequently with their advisors at the early stage of their careers. Similarly, we use \bm{$A_{e}$} $\in \mathbb{R}^{{{n_2} \times m_{2}}}$ to preserve the edge attributes, where $n_{2} = |E|$ is the number of edges and edge attribute categories is defined as $m_{2}$.
\end{myDef}
TABLE~\ref{tab:1} lists other symbols used in the representation learning process.

\begin{table}[h!]
\centering
\caption{Description of notations}
\label{tab:1}
\begin{tabular}{m{2cm}<{\centering} |m{6cm}<{\centering} } 
\hline
\hline
Notation & Description \\
\hline
\hline
$n_{1} = |V|$ & the number of nodes in the collaboration network $G$\\
$n_{2} = |E|$ & the number of edges in the collaboration network $G$\\
$L $& the label set for edges\\
$m_{1}$ & the number of node attribute categories \\
$m_{2}$ & the number of edge attribute categories \\
$d_{1}$ & the dimension of node representation\\
$d_{2}$ & the dimension of edge representation\\
$r$ & the learning rate for the autoencoder\\
$\bm{A_{n}}\in \mathbb{R}^{{{n_1} \times m_{1}}}$ & the node attribute information matrix\\
$\bm{A_{e}}\in \mathbb{R}^{{{n_2} \times m_{2}}}$ & the edge attribute information matrix\\
$\bm{H} \in \mathbb{R}^{{{n_1} \times d}}$ & the final representation of the network\\
\hline
\end{tabular}
\end{table}%

Based on the notations explained above, the problem of advisor-advisee relationship mining can be described as follows:

\textbf{Input}: A scholar $i$ who is associated with \bm{$A_{n}$} and \bm{$A_{e}$}, the collaboration network $G$, and $C_i$.

\textbf{Output}: Who is $i$'s advisor?

\subsection{Framework of Shifu2}
We assume both nodes and edges attributes can preserve the interactions between advisors and advisees in a collaboration network. From the perspective of node attributes, advisors usually publish more papers than their advisees. Meanwhile, advisors will have a relatively long academic career, which leads to cumulative advantages, both in academic performance and resources. On the other hand, for edges attributes, the collaboration mechanism between advisors and advisees is obviously different from others. From the perspective of advisees, they collaborate more closely with their advisors. So we need to model the attribute proximity for both nodes and edges.

With this hypothesis, the objective of the collaboration network re-construction part in Shifu2 is to minimize the construction loss as:
\begin{equation}
\label{eq1}
L_{sf} = L_{a} + L_{e}
\end{equation}
where $L_{a}$ and $L_{e}$ represent the reconstructed loss for the processes of node representation learning and edge representation learning. Followings give details about each part.

%
%
%
%
%
\subsubsection{\textbf{Node Representation Learning}}
Nodes attributes are highly correlated with the network structure in social networks~\cite{marsden1988homogeneity}. In order to preserve the proximity of each scholar from the perspective of nodes connection, motivated by the method proposed by Tu et al.~\cite{tu2017transnet}, we adopt a deep autoencoder to get the embedding representation \bm{$A'$}. The hidden layers of the autoencoder are considered as two parts: the encoder part and the decoder part. The layers consistently encode and decode the input data. The output of the $(i-1)th$ layer is considered as the input of the $i$th layer, which can ensure that the output is equal to the input. The hidden layers can automatically capture the characteristics of input data and keep them unchanged.

For each scholar and his/her collaborators in the collaboration network, we use the adjacency matrix represented by $\bm{A}$ as the input of the node autoencoder, where $\bm{A}_{ij} =1$ if $(v_i, v_j)\in E$ and $\bm{A}_{ij} = 0$, otherwise. Meanwhile, we consider the attributes listed in TABLE~\ref{tab:2} for each node in the network.

\begin{table}[htbp]
\centering
\caption{Description of node features}
\label{tab:2}
\begin{tabular}{m{1cm}<{\centering} |m{7cm}<{\centering}} 
\hline
\hline
Notation & Description \\
\hline
\hline
$aa_i$ & the academic age of the scholar $i$\\
$aa_j$ & the academic age of the scholar $j$\\
$org_i$ & the organization of the scholar $i$\\
$org_j$ & the organization of the scholar $j$\\
$np_{i}$ & the number of publications of $i$ before collaborating with $j$ \\
$np_{j}$ & the number of publications of $j$ before collaborating with $i$ \\
\hline
\end{tabular}
\end{table}

$aa$ represents a scholar's academic age when he/she began to collaborate with his/her collaborators. It can be calculated as~(\ref{eq2}):
\begin{equation}
\label{eq2}
  aa = y_{c} -y_{f}
\end{equation}
where $y_{f}$ is the year when $i$ published the first paper and $y_{c}$ is the year when $i$ first co-authored with his/her collaborators. For example, if scholar $i$ published his/her first paper in 1987 and the first co-authored paper with $j$ is published in 2000, then his $aa$ when he first collaborated with $j$ is 13 (2000-1987 = 13).
%

In each hidden layer, we adopt the following no-linear transformation function:
\begin{equation}
\begin{aligned}
\label{eq3}
   &\bm{h}_{(1)}^{a} = f_{a} (\bm{W}_{(1)}^{a}\bm{x} + \bm{b}_{(1)}^{a}) \\
   &\bm{h}_{(i)}^{a} = f_{a}(\bm{W}_{(i)}^{a}\bm{h}_{(i-1)}^{a} + \bm{b}_{(i)}^{a}), i = 2,...,k
\end{aligned}
\end{equation}
where $f_{a}$ is the activation function and $\bm{W}_{(i)}^{a}$, $\bm{b}_{(i)}^{a}$ represent the transformation matrix and the bias vector, respectively. $k$ is the total number of hidden layers. Besides, we utilize the Sigmoid function to map vectors of arbitrary real values to the range $[0,1]$.

The goal of the node representation part is to minimize the reconstruction error between the representation vectors and the original input. However, the number of non-zero elements in the input is far less than that of zero elements. Therefore, it is easy to reconstruct zero elements. To address this problem, we impose more penalties on reconstruction error of non-zero elements. According to Wang et al.~\cite{wang2016structural}, we add a non-zero penalty matrix $\bm{A''}$ and define the objective function as:
\begin{equation}
\begin{aligned}
\label{eq5}
    L_{a} = ||((\bm{A_{n}\parallel A}) - \bm{A'})\odot \bm{A''}||_{F}^{2}
\end{aligned}
\end{equation}
where $\bm{A_{n}\parallel A}$ is the concatenation operation of $\bm{A_{n}}$ and $\bm{A}$. Specifically, the concatenation operation $\parallel$ in this equation is to connect two matrices in the horizontal axis. Here we use a toy example to explain this operation clearly. Suppose that there are three nodes $n_1,n_2,n_3$ in the network $G$ nd the adjacency matrix $\bm{A} = [[0,1,0],[1,0,1],[0,1,0]]$. Each node in G is associated with two attributes and the node attributes matrix is $\bm{A_{n}} = [[2,1],[1,0],[1,1]]$. Then $\bm{A_{n}\parallel A} = [[0,1,0,2,1],[1,0,1,1,0],[0,1,0,1,1]]$. The concatenation operation is implemented by the function $numpy.concatenate((), axis=1)$ in Python. If $\bm{A}_{ij} > 0$, then $\bm{A''}_{ij} = \rho,  \rho>1$, else $\bm{A''}_{ij} = 1$. By using the loss function as~(\ref{eq5}), the nodes with similar characteristics will be close in the embedding space.

We adopt Adaptive moment estimation (Adam)~\cite{DBLP:journals/corr/KingmaB14} method for first-order gradient-based optimization. For each step in the iteration process, we need to compute the gradients $g_{t}$ at time $t$. Then we compute the first moment estimate $m_t$ and the second raw moment estimate $v_t$, where $m_{t}$ is the mean of the gradient $g$ and $v_{t}$ is the non-central variance of $g_t$. According to the bias-corrected $\widehat{m}_{t}$ and $\widehat{v}_{t}$, we can update the parameter $\theta_t $. In summary, the optimization process can be described as:
\begin{equation}
\begin{aligned}
\label{eq6}
g_t & \longleftarrow \bigtriangledown_\theta f_t( \theta_{t-1})\\
m_t & \longleftarrow \beta_{1} \cdot m_{t-1} + (1-\beta_{1}) \cdot g_t \\
v_t & \longleftarrow \beta_{2} \cdot v_{t-1} + (1-\beta_{2}) \cdot g_t^2 \\
\widehat{m}_{t} &\longleftarrow m_{t} / (1-\beta_1^t)\\
\widehat{v}_{t} &\longleftarrow v_{t} / (1-\beta_2^t)\\
\theta_t & \longleftarrow \theta_{t-1}-\alpha \cdot \widehat{m}_{t}/(\sqrt{\widehat{v}_{t}}+ \epsilon).\\
\end{aligned}
\end{equation}
Based on experience and experimental results, in our optimization process, we set 0.9 for $\beta_{1}$, 0.999 for $\beta_{2}$, and $10^{-8}$ for $\epsilon$.

\subsubsection{\textbf{Edge Representation Learning}}
In the edge representation construction process, we also employ the deep autoencoder to convert the attribute matrix to the low-dimensional vector representations. The reconstruction process and the implementation details are presented below.

In order to preserve the edge proximity, we take the edge attribute matrix as the input of edge autoencoder. We consider attribute information listed in TABLE~\ref{tab:3} for each edge in the network.

\begin{table}[htbp]
\centering
\caption{Description of edge features}
\label{tab:3}
\begin{tabular}{m{1cm}<{\centering} |m{6.8cm}<{\centering}} 
\hline
\hline
Notation & Description \\
\hline
\hline
$ad_{ij}$ & difference of academic age between $i$ and $j$\\
$ct_{ij}$ & the collaboration times of $i$ and $j$\\
$cd_{ij}$ & the collaboration duration of $i$ and $j$\\
$ft$ &the number of times $i$ and $j$ being the first two authors\\
$lf$ &the number of times $i$ and $j$ being the first and the last authors\\
$kulc^{t}_{ij}$ & the collaboration similarity between $i$ and $j$  \\
\hline
\end{tabular}
\end{table}
$kulc_{y}^{ij}$ is the collaboration similarity between scholar $i$ and his/her collaborator $j$ after $t$ years since they first co-authored a paper~\cite{wu2010re}. It aims to depict the dynamics of collaboration. We compute $kulc$ as:
\begin{equation}
\label{eq7}
  kulc^{t}_{ij} = \frac{np_{ij}}{2} (\frac{1}{np_{i}} + \frac{1}{np_{j}})
\end{equation}
where $np_{ij}$ is the number of co-authored papers in $t$ years, and $np_{i}$, $np_{j}$ represent the number of publications of $i$, $j$, respectively.

The autoencoder adopts the edge attribute matrix as input. It encodes and decodes the vector in each no-linear transformation layers as:
\begin{equation}
\begin{aligned}
\label{eq8}
   &\bm{h}_{(1)}^{e} = f_{e} (\bm{W}_{(1)}^{e}a + \bm{b}_{(1)}^{e}) \\
   &\bm{h}_{(j)}^{e} = f_{e}(\bm{W}_{(j)}^{e}\bm{h}_{(j-1)}^{e} + \bm{b}_{(j)}^{e}), j = 2,...,m
\end{aligned}
\end{equation}
$f_{e}$ is the activation function and $m$ is the number of hidden layers of the encoder. In the $j$th layer, we use $\bm{W}_{(j)}^{e}$ and $\bm{b}_{(j)}^{e}$ to represent the transformation matrix and the bias vector, respectively. Specifically, the input of the edge autoencoder is dense, so we utilize the dropout~\cite{srivastava2014dropout} to overcome the overfitting. Similar to the process of nodes representation, Sigmoid function is employed as the activation function to get the edge representation matrix $\bm{E}'$. The reconstructed loss is computed as:
\begin{equation}
\begin{aligned}
\label{eq9}
    L_{e} = ||\bm{{A}_{e}} - \bm{E}'||_{F}^{2}.
\end{aligned}
\end{equation}
Hence, we can minimize the distances between the reconstructed representations and the original input.

\subsubsection{\textbf{Pooling Layer}}
Some fields contain thousands of scholars. It is difficult to ensure the computational time and memory for the running control with numerous vectors. As a result, we add a pooling layer to compress input features. We first reduce each adjacency vector to 1000 dimensions and calculate the mean of reduced vectors accordingly as the input of encoders.

\subsubsection{\textbf{Advisor-advisee Relationship Identification}}
After the processes of nodes and edges reconstruction, we can obtain the low-dimensional representations of the collaboration network. Then we need to use a supervised classifier to predict the label for each edge.

More specifically, for the unlabeled edge $e = (i,j)$, Shifu2 aims to identify the label for $e$. To achieve this goal, we add a supervised classifier in our Shifu2 model, which takes the output of the last hidden layer as input and returns the classification results. We adopt logistic regression~\cite{harrell2015ordinal} as the classification method in our model. Mathematically, the loss function $L_{lr}$ is computed as:
\begin{equation}
\begin{aligned}
\label{eq10}
    L_{lr} = |L - (\bm{D} \cdot \bm{W}^{l} + \bm{B}^{l})|
\end{aligned}
\end{equation}
where $\bm{D}$ is the concatenation of two encoders. $\bm{W}^{l}$ is the weight of logistic regression and $\bm{B}^{l}$ is the bias. In this part, we also use Adam method for stochastic optimization.

\subsubsection{\textbf{Overall Reconstruction}}
After the representation process from the perspectives of both nodes and edges, in order to preserve nodes and edges properties at the same time, we calculate the joint loss function to optimize the objective as:
\begin{equation}
\begin{aligned}
\label{eq11}
    L_{sum} = &L_{sf}  + \beta L_{lr} + \\
              &\alpha (\Sigma_{i} (\bm{W}^{e}_{i} + \bm{B}^{e}_{i}) +  \gamma \Sigma_{j} (\bm{W}^{a}_{j}+ \bm{B}^{a}_{j}))
\end{aligned}
\end{equation}
where $\alpha $ and  $\beta$ are two hyperparameters to adjust the weights of the edge encoder and the node encoder.

Since the edge encoder is a process of dimension rise and the node encoder is a process of dimension reduction, the sum of weight and bias of the node encoder
is much larger than the edge side. We use the regularization parameter $\gamma$ to balance in the middle, to prevent the value of the node side overweight the edge side. In the experiment, we set $\gamma$ as 0.01. If we combine~(\ref{eq11}) with~(\ref{eq1}),~(\ref{eq5}),~(\ref{eq9}) and~(\ref{eq10}), then the overall loss function can be rewritten as:
\begin{equation}
\begin{aligned}
\label{eq12}
    L_{sum} = &||((\bm{A_{n}\parallel A}) - \bm{A'})\odot \bm{A''}||_{F}^{2} + ||\bm{A_{e}} - \bm{E'}||_{F}^{2} + \\
              & \beta |L - (\bm{D} \cdot \bm{W}^{l} + \bm{B}^{l})| +\\
              & \alpha (\Sigma_{i} (\bm{W}^{e}_{i} + \bm{B}^{e}_{i}) + \gamma\Sigma_{j} (\bm{W}^{a}_{j} + \bm{B}^{a}_{j})).
\end{aligned}
\end{equation}

The goal of the aforementioned model is to minimize the loss function $L_{sum}$. In this paper, we achieve this goal by using back propagation algorithm (BP) with stochastic gradient descent, e.g., $W_{ji} = W_{ji}- \delta\frac{\partial}{\partial W_{ji}}L(X,Y)$. If $z = Wx + b$, the gradient can be defined as:
\begin{equation}
\begin{aligned}
\label{eq13}
\frac{\partial L(X,Y)}{\partial W_{ji}} &= \Sigma^{n}_{i=1} \frac{\partial L(X,Y)}{\partial Z_{j}} \cdot \frac{\partial Z_{j}}{\partial W_{ji}} = \Sigma^{n}_{i=1} \rho_{j} x_{i}^{T}\\
\frac{\partial L(X,Y)}{\partial b_{j}} &= \Sigma^{n}_{i=1} \frac{\partial L(X,Y)}{\partial Z_{j}} \cdot \frac{\partial Z_{j}}{\partial b_{j}}=  \Sigma^{n}_{i=1} \rho_{j}\\
\end{aligned}
\end{equation}
where $\rho_{j}= \partial L(X,Y) / \partial z_{j}$  is the reconstruction error between the activation and the target. $n$ represents the number of samples.

If $f'(.)$ represents the partial derivative of $f(.)$, then the updated error can be summarized as:
\begin{equation}
\begin{aligned}
\label{eq14}
\rho_{j}^{(Y)} &= -\Sigma_{i=1}^{n}(y_{ij}-h_{ij}) \cdot f'(z_{j}^{(Y)})\\
\rho_{j}^{(H)} &= \Sigma_{i=1}^{n}W_{ji}^{H} \rho_{i}^{(Y)}\cdot f'(z_{j}^{(H)}).
\end{aligned}
\end{equation}

\begin{figure}
  \centering
  \includegraphics[width=0.5\textwidth]{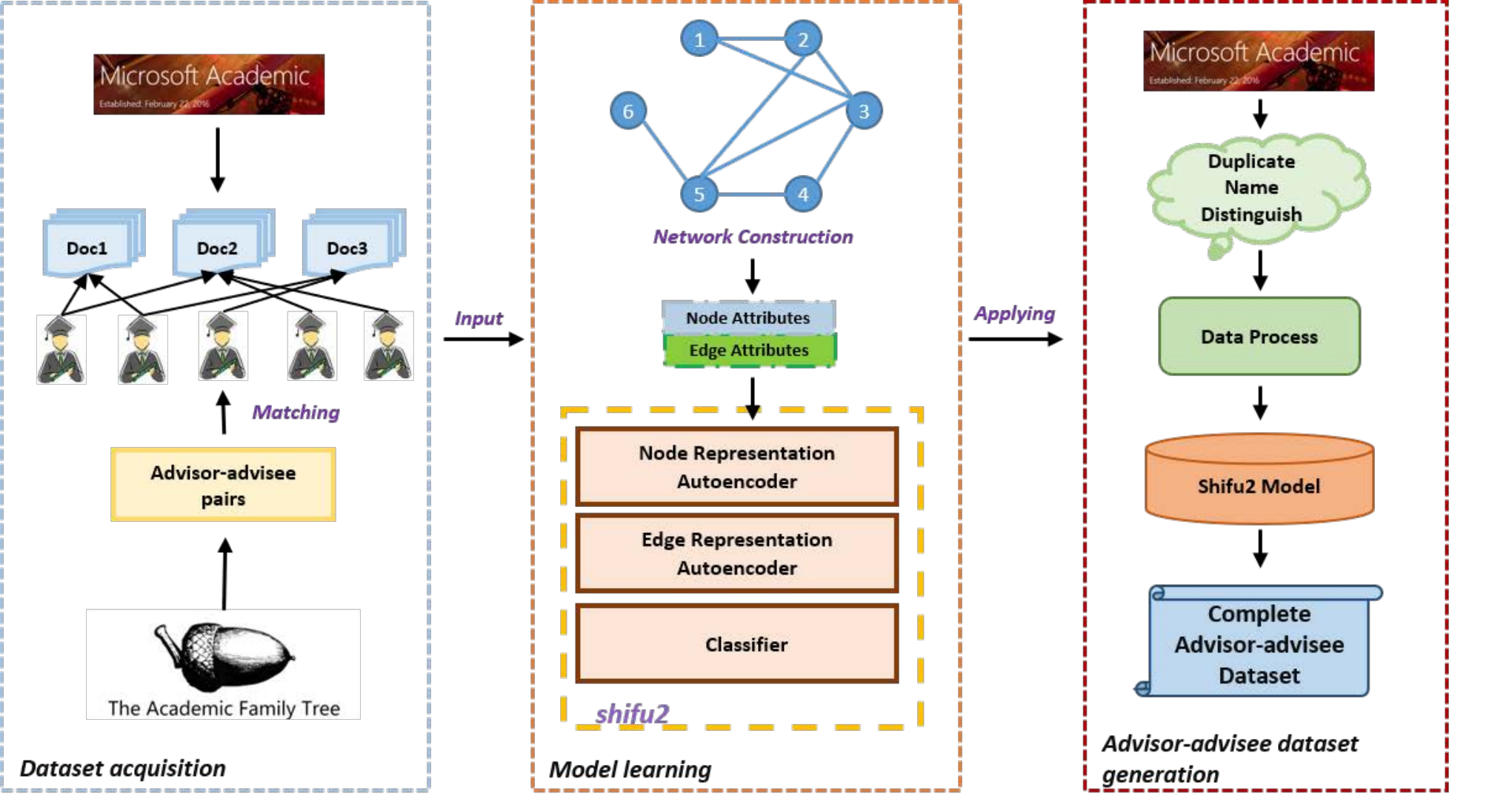}\\
  \caption{Experiment procedures.}
  \label{fig:3}
\end{figure}

The overall architecture of the algorithm is summarized in Algorithm 1.

\begin{algorithm}
\label{alg1}
\caption{The advisor-advisee relationship mining algorithm.}
    \begin{algorithmic}[1] 
    \Require $\bm{A}$, \bm{$A_{n}$}, \bm{$A_{e}$}, $\alpha, \beta, \gamma$, $r$, $L$, and the convergence condition $\epsilon$.
    \Ensure $l$ for each $e = (i,j)$\\
    \textbf{Initiate}: randomly initiate $\theta_{1} \leftarrow \{\bm{W}^{e}_{i}, \bm{B}^{e}_{i}\}_{i=1}^{m}$,  $\theta_{2} \leftarrow \{\bm{W}^{a}_{j}, \bm{B}^{a}_{j}\}_{j=1}^{n}$ , and  $\theta_{3} \leftarrow \{\bm{W}^{l}, \bm{B}^{l}\}$ .\\
    \textbf{Pre-train}: set $r = 0.01$, train edge auto-encoder and node auto-encoder independently.\\
    \textbf{Train}:
        \While{$\epsilon$ is true}
                \State{randomly generate a batch of data from advisor-advisee edge data and advisors and advisee's attributes}
                \If{$\bm{A}_{ij} > 0 $}
                    \State {$\bm{A''}_{ij} \leftarrow \rho$ , $\rho >1 $}
                \Else
                    \State { $\bm{A''}_{ij} \leftarrow 1$}
                \EndIf
                \State{calculate $L_{a}$ based on~(\ref{eq5}), get reconstructed \bm{$A'$}}
                \State{calculate $L_{e}$ based on~(\ref{eq9}), get reconstructed \bm{$E'$}}
                \State{concatenate $\bm{D} \leftarrow [\bm{A'}, \bm{E'}]$ }
                \State{calculate $L_{sum}$ based on~(\ref{eq12})}
                \State{update $\theta_{1}, \theta_{2}, \theta_{3}$ }
        \EndWhile
    \end{algorithmic}
\end{algorithm}

\subsection{Algorithm analysis}
In the proposed model, we first generate \bm{$A'$} and \bm{$E'$} through the node autoencoder and the edge autoencoder. The complexity of generating the above embedding matrix is $O(c_en_ed_eI_e+c_an_ad_aI_a)$, where $c_e$, $c_a$ are the size of samples. $n_e$, $n_a$ are the dimension of the attributes. $d_e$, $d_a$ are the maximum dimension of the hidden layer, and $I_e$, $I_a$ are iterations times. For the logistic regression process, the computational complexity is $O(n*k + k)$, where $n$ is the sample size and $k$ represents the feature dimension, which is related to the embedding dimension of the encoders. It is not difficult to see that the total training complexity of Shifu2 is $O(c_en_ed_eI_e+c_an_ad_aI_a+nk+k)$. Approximately, the time complexity of Shifu2 is $O(cndI)$.


\section{Experiments}
\label{sec4}
This section presents the experimental details to assess the accuracy and the effectiveness of the proposed model. Fig.~\ref{fig:3} describes the complete process of the task. Below we will give the details of each procedure.

\begin{figure}
  \centering
  \includegraphics[width=0.4\textwidth]{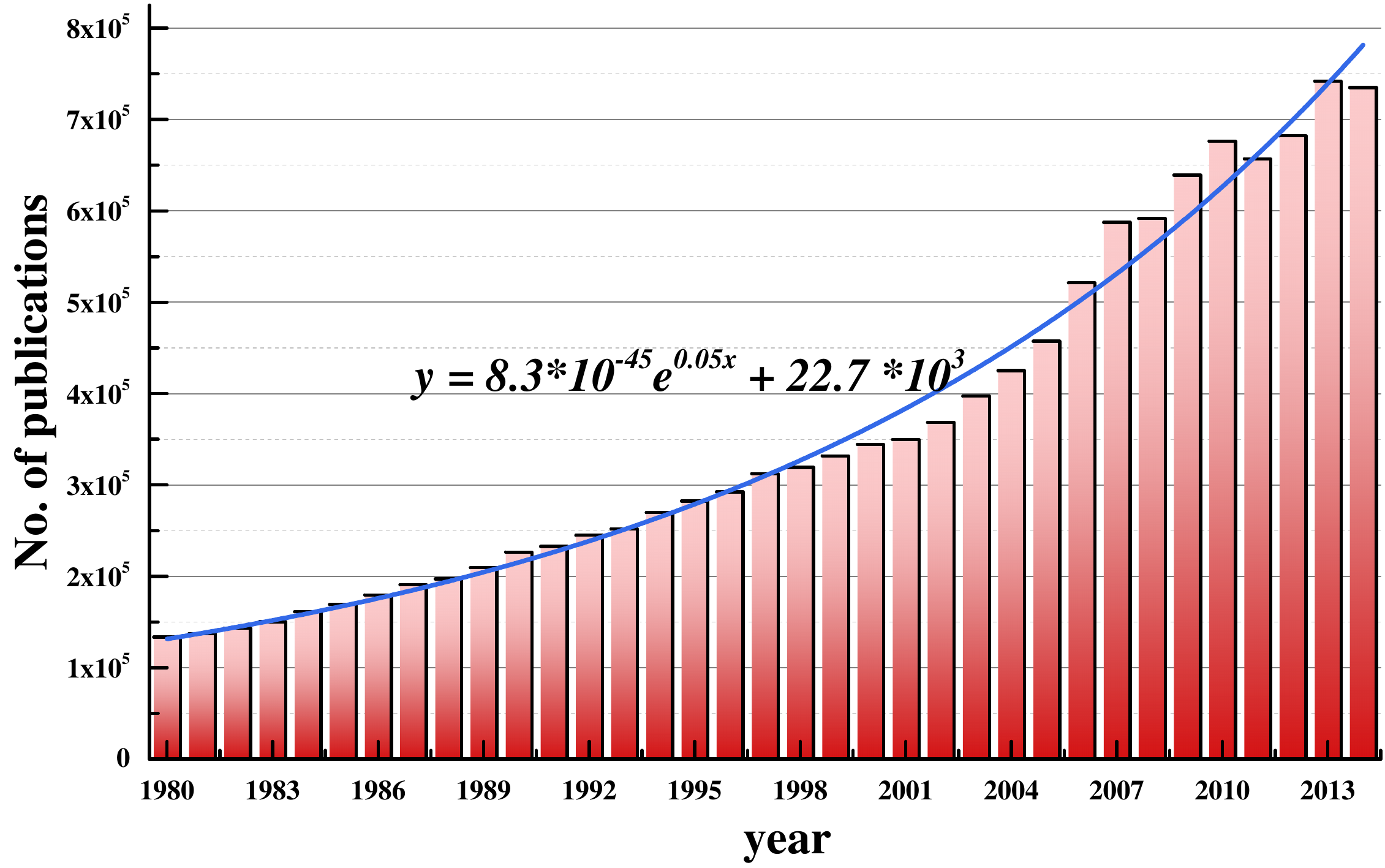}\\
  \caption{Number of new records per year in the MAG dataset. }
  \label{fig:4}
\end{figure}

\subsection{Experimental Setup}
\subsubsection{\textbf{Datasets}}
\label{sec53}
As mentioned previously, we use the MAG which is a widely used dataset containing scientific publication records to construct the collaboration network. In order to investigate the accuracy and the effectiveness of Shifu2, we need some ground truth advisor-advisee pairs. We extract the ground truth advisor-advisee pairs from The Academic Family Tree (AFT)\footnote{https://academictree.org/}, which is a user content-driven web database storing academic genealogy. We crawl the realistic advisor-advisee pairs in the fields of Chemistry, Computer Science, Economics, Engineering, Mathematics, and Physics. TABLE~\ref{tab:4} lists the statistics for advisor-advisee pairs in major research fields.
\begin{table}[htbp]
\centering
\caption{Statistics of advisor-advisee pairs in each field}
\label{tab:4}
\begin{tabular}{|m{2.3cm}<{\centering} |m{3.3cm}<{\centering}|m{1.5cm}<{\centering}|} 
\hline
Field& Number of advisor-advisee pairs& Time period \\
\hline
Chemistry &28,449&2000-2010\\
Computer Science&8853&2000-2010\\
Economics&2666&2000-2010\\
Engineering&9176&2000-2010\\
Mathematics& 5962&2000-2010\\
Physics&7654&2000-2010\\
\hline
\hline
Total&\multicolumn{2}{c|}{62,760} \\
\hline
\end{tabular}
\end{table}

Based on the ground truth advisor-advisee dataset, which only partially covers the authors in the MAG, we further randomly separate the dataset into two sub-datasets: the training set and the test set. The training set contains advisor-advisee pairs who co-authored the first paper during 2000-2006, and the others are used as the test set.

\begin{figure*}[!ht]
\centering
\subfigure[Chemistry]{
\label{fig:5-a}
\includegraphics[width=0.425\textwidth]{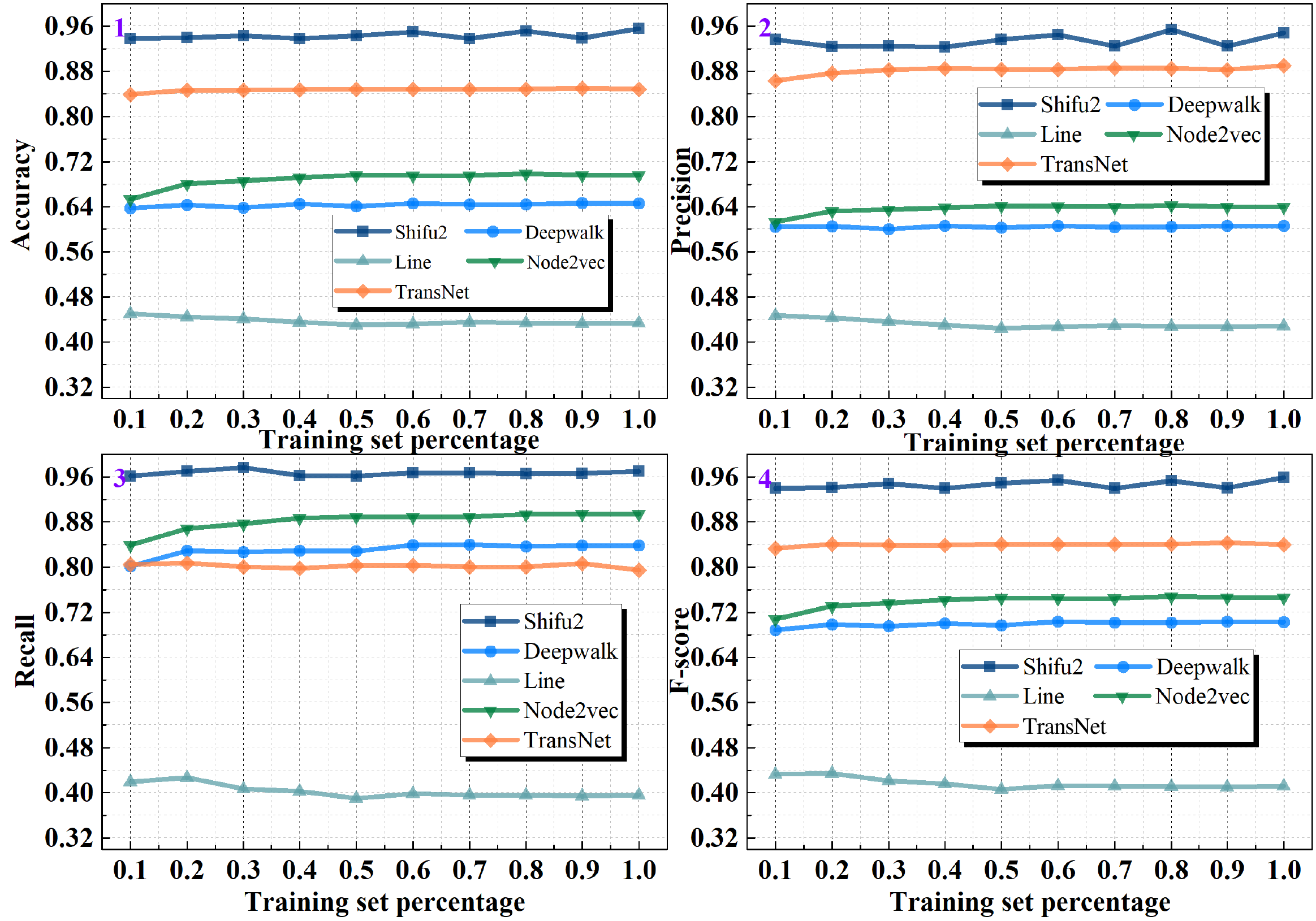}}
\subfigure[Computer Science]{
\label{fig:5-b}
\includegraphics[width=0.425\textwidth]{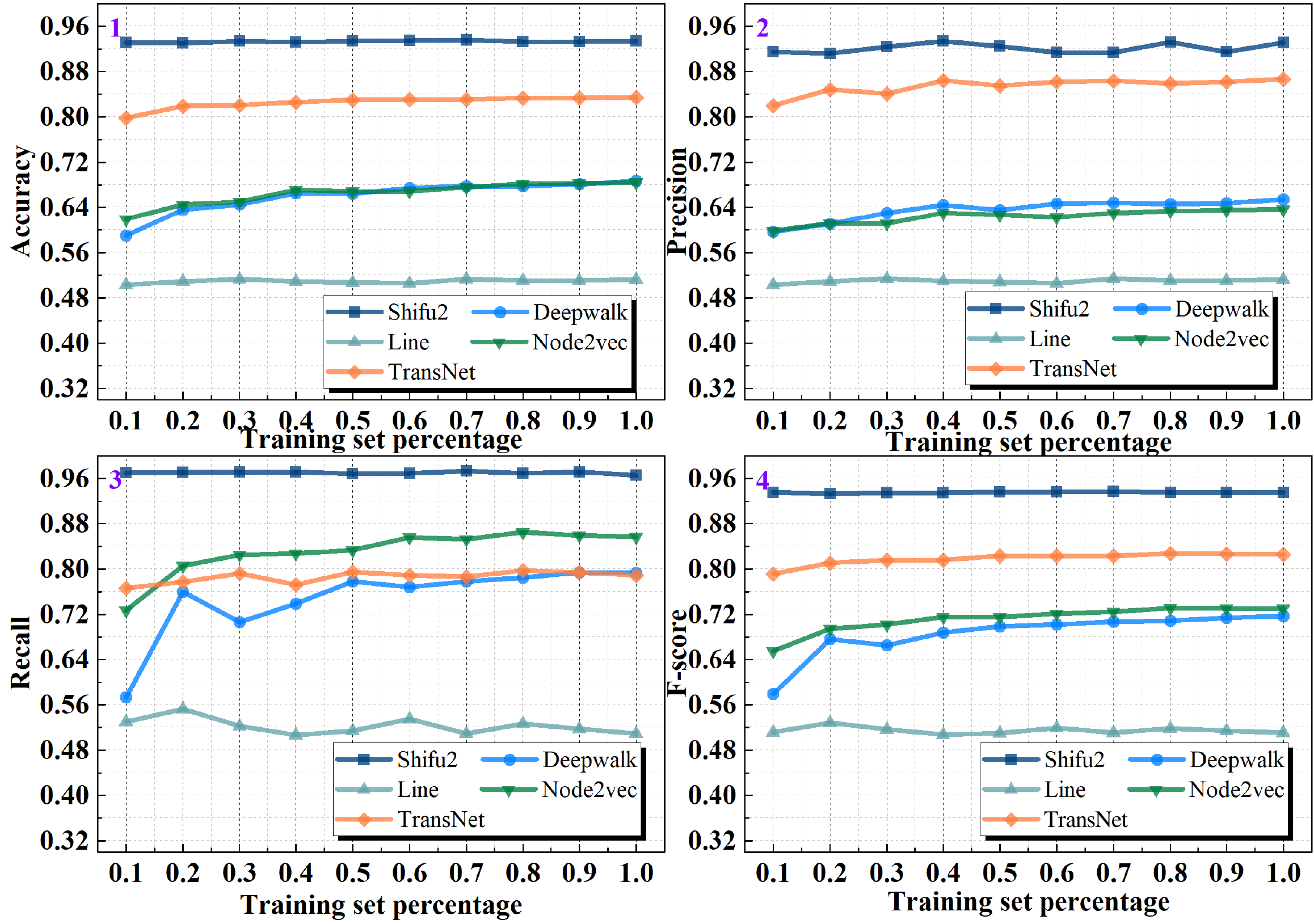}}\\
\subfigure[Economics]{
\label{fig:5-a}
\includegraphics[width=0.425\textwidth]{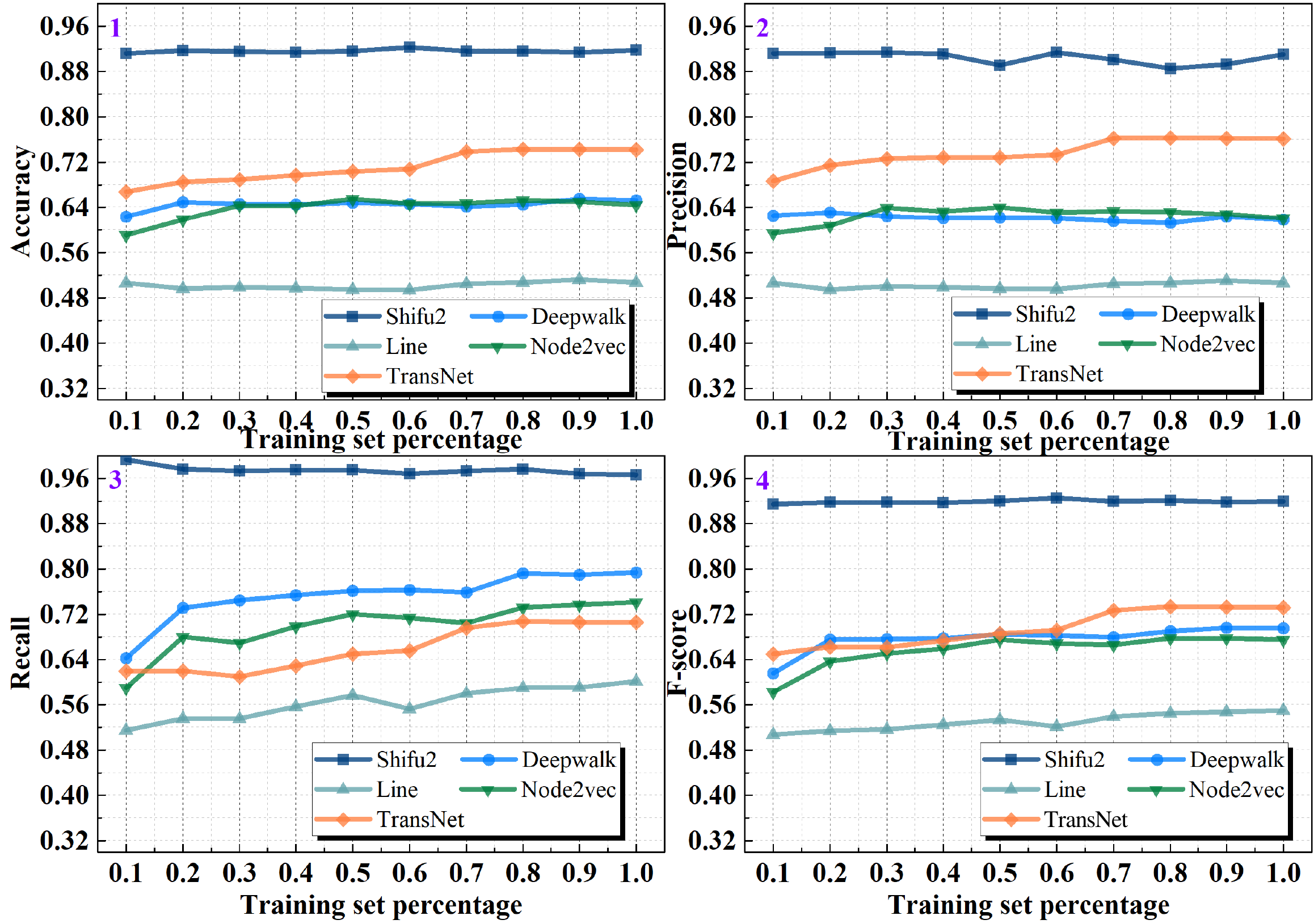}}
\subfigure[Engineering]{
\label{fig:5-a}
\includegraphics[width=0.425\textwidth]{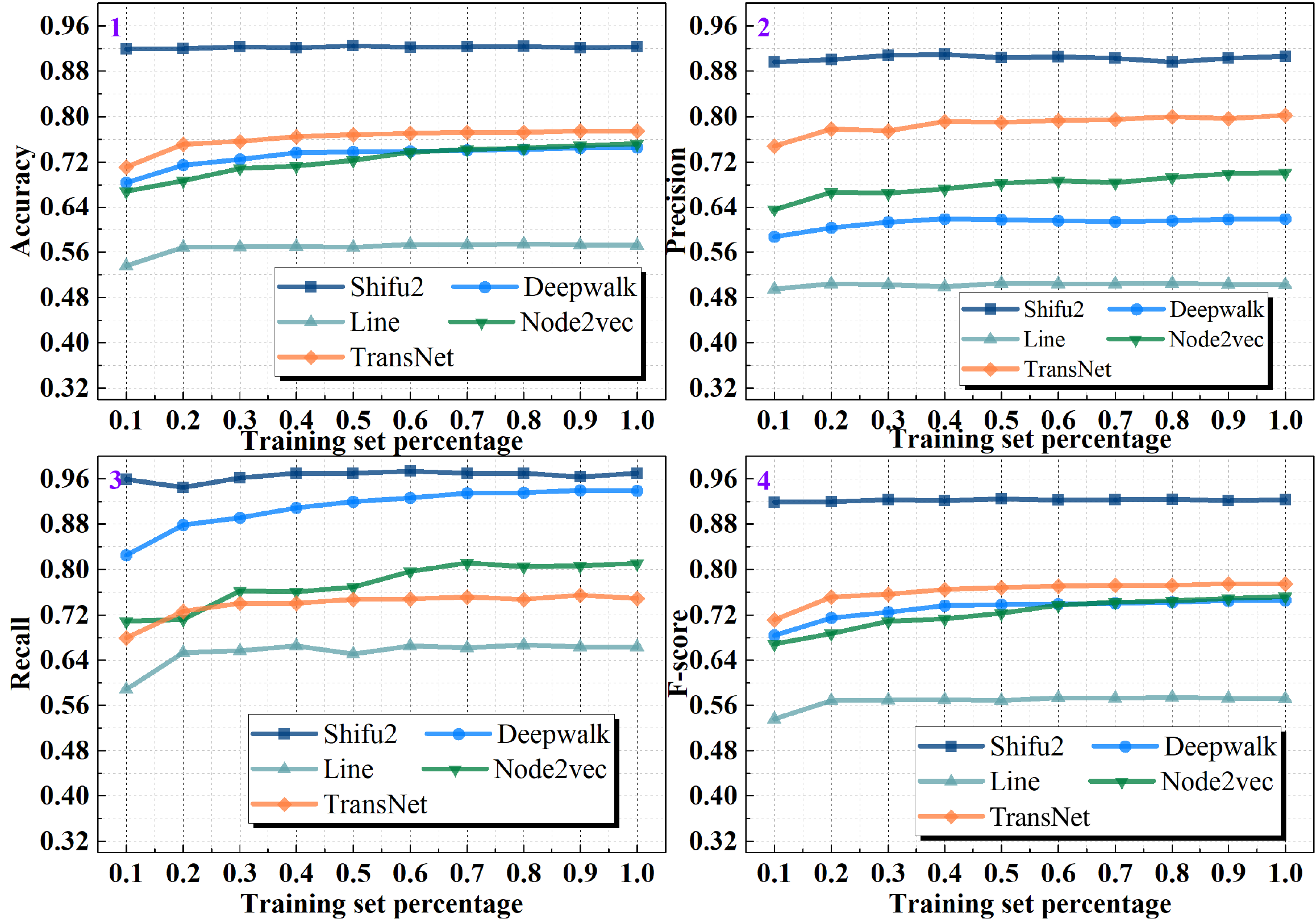}}\\
\subfigure[Mathematics]{
\label{fig:5-a}
\includegraphics[width=0.425\textwidth]{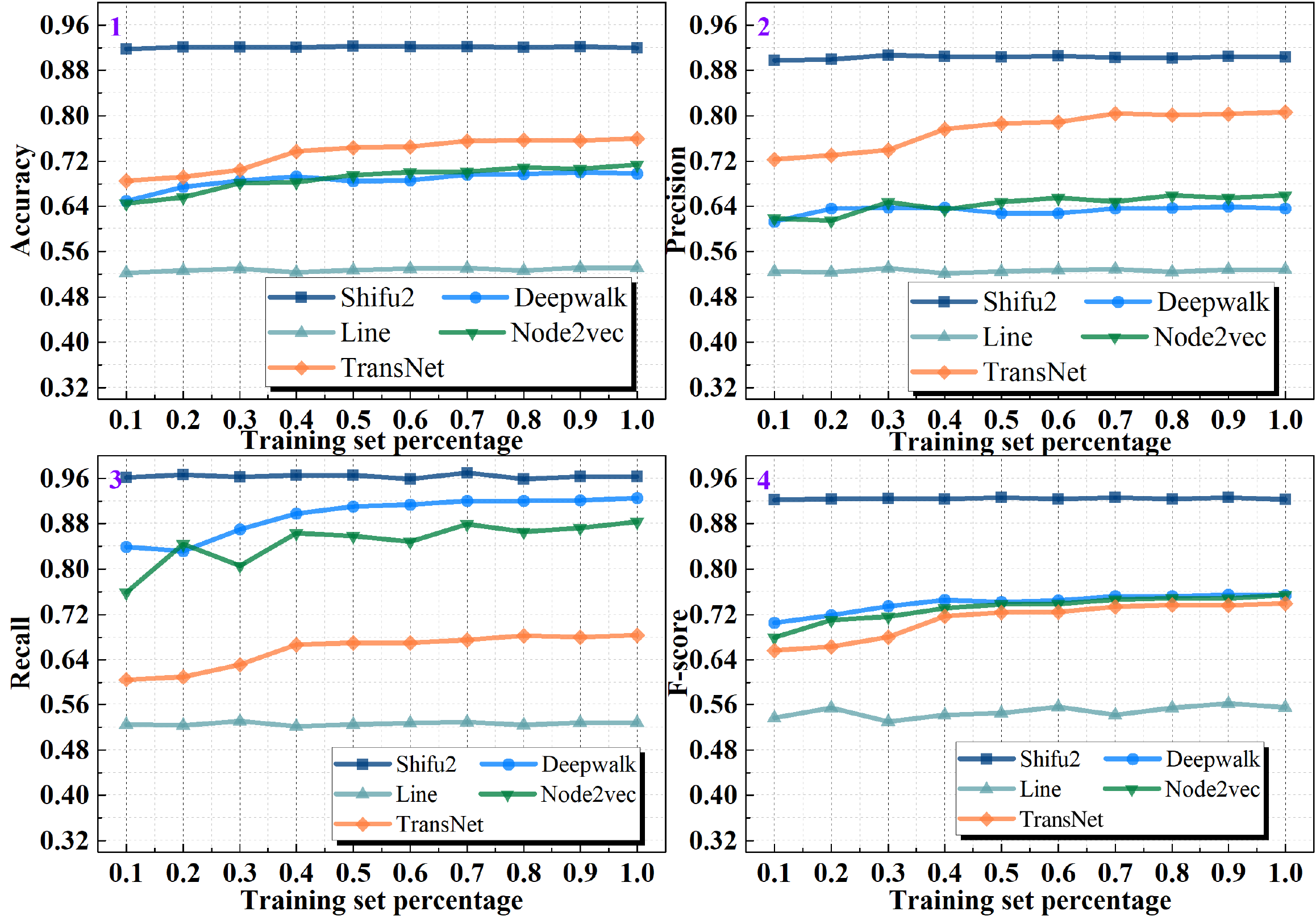}}
\subfigure[Physics]{
\label{fig:5-a}
\includegraphics[width=0.425\textwidth]{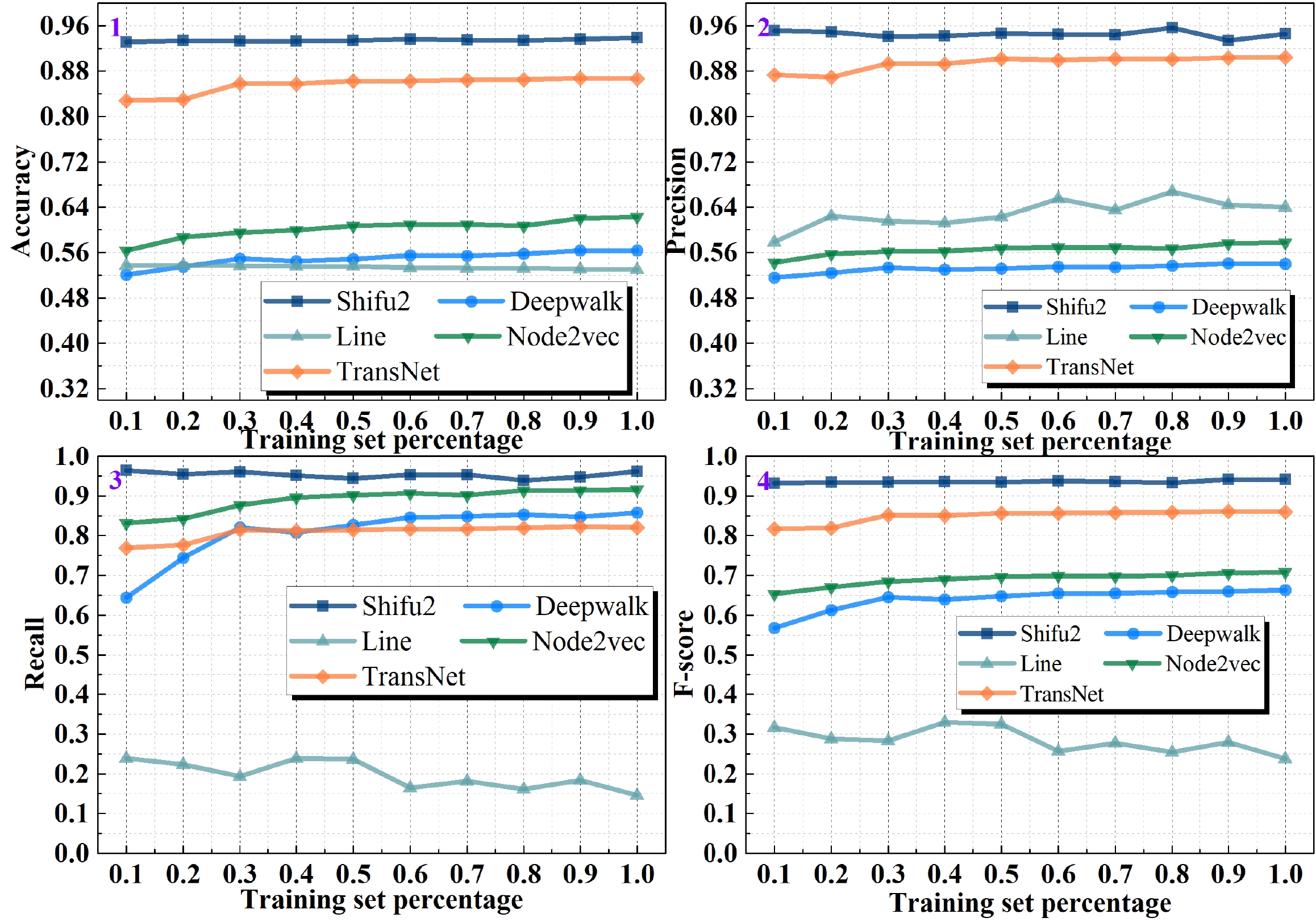}}\\
\caption{Performance of different methods with different sizes of training data. }
\label{fig:5}
\end{figure*}
\subsubsection{\textbf{Datasets Pre-processing}}
Before applying Shifu2 onto the entire MAG dataset to generate the academic genealogy, there are a set of challenges that must be addressed:
\begin{itemize}
  \item \textbf{Author name disambiguation}. The MAG only provides the publication related information such as title, authors, published year, and so on. Authors' names have not been pre-processed which means that if two scholars have the same name, their publications cannot be distinguished. Duplicated scholars' personal information challenges the application of the model.
  \item \textbf{Disciplinary differences}. It is now well established from a variety of studies, that diverse collaborative ties exist in different research fields~\cite{iglivc2017whom,montoya2018fast}. The disciplinary differences may affect the performance of the model. It is a challenge that how to eliminate the disciplinary differences and make the model universal.
  \item \textbf{Temporal effect}. Fig.~\ref{fig:4} illustrates that the number of publications in the MAG steadily grows over time, which may introduce influence on the law of advisor-advisee's collaboration. It is important to make sure that the results are not affected by the temporal effect.
\end{itemize}

In order to address these challenges, we pre-process the dataset as follows:
\begin{table*}[htbp]
  \centering
  \caption{Advisor-advisee relationship identification performance of different methods}
    \begin{tabular}{c|c|c|c|c|c|c|c|c|c|c}
    \toprule
          & \multicolumn{5}{c|}{Chemistry}        & \multicolumn{5}{c}{Computer Science} \\
    \midrule
     \diagbox{Metrics}{Method} & Shifu2 & Deepwalk & Line  & Node2vec & TransNet & Shifu2 & Deepwalk & Line  & Node2vec & TransNet \\
    \midrule
    Accuracy & \textcolor[rgb]{ 1,  0,  0}{\textbf{0.939}} & 0.646 & 0.451 & 0.699 & 0.850 & \textcolor[rgb]{ 1,  0,  0}{\textbf{0.931}} & 0.681 & 0.514 & 0.685 & 0.834\\
    \midrule
    Precision & \textcolor[rgb]{ 1,  0,  0}{\textbf{0.925}} & 0.606 & 0.447 & 0.643 & 0.890 & \textcolor[rgb]{ 1,  0,  0}{\textbf{0.912}} & 0.655 & 0.514 & 0.637 & 0.867\\
    \midrule
    Recall & \textcolor[rgb]{ 1,  0,  0}{\textbf{0.958}} & 0.840 & 0.427 & 0.894 & 0.808& \textcolor[rgb]{ 1,  0,  0}{\textbf{0.959}} & 0.794 & 0.552 & 0.865 & 0.797 \\
    \midrule
    F1-score & \textcolor[rgb]{ 1,  0,  0}{\textbf{0.941}} & 0.704 & 0.427 & 0.748 & 0.843& \textcolor[rgb]{ 1,  0,  0}{\textbf{0.933}} & 0.717 & 0.528 & 0.731 & 0.827 \\
    \midrule
          & \multicolumn{5}{c|}{Economics}        & \multicolumn{5}{c}{Engineering} \\
    \midrule
     \diagbox{Metrics}{Method}  & Shifu2& Deepwalk & Line  & Node2vec & TransNet & Shifu2 & Deepwalk & Line  & Node2vec & TransNet \\
    \midrule
    Accuracy & \textcolor[rgb]{ 1,  0,  0}{\textbf{0.913}} & 0.656 & 0.512 & 0.656 & 0.743 & \textcolor[rgb]{ 1,  0,  0}{\textbf{0.915}} & 0.680 & 0.506 & 0.733 & 0.782\\
    \midrule
    Precision & \textcolor[rgb]{ 1,  0,  0}{\textbf{0.877}} & 0.631 & 0.507 & 0.641 & 0.763 & \textcolor[rgb]{ 1,  0,  0}{\textbf{0.889}} & 0.619 & 0.505 & 0.702 & 0.802 \\
    \midrule
    Recall & \textcolor[rgb]{ 1,  0,  0}{\textbf{0.961}} & 0.794 & 0.602 & 0.741 & 0.708 & \textcolor[rgb]{ 1,  0,  0}{\textbf{0.952}} & 0.940 & 0.667 & 0.812 & 0.755 \\
    \midrule
    F1-score & \textcolor[rgb]{ 1,  0,  0}{\textbf{0.917}} & 0.697 & 0.550 & 0.678 & 0.734 & \textcolor[rgb]{ 1,  0,  0}{\textbf{0.919}} & 0.746 & 0.575 & 0.753 & 0.891 \\
    \midrule
          & \multicolumn{5}{c|}{Mathematics}       & \multicolumn{5}{c}{Physics} \\
    \midrule
     \diagbox{Metrics}{Method}  & Shifu2 & Deepwalk & Line  & Node2vec & TransNet & Shifu2 & Deepwalk & Line  & Node2vec & TransNet \\
    \midrule
    Accuracy & \textcolor[rgb]{ 1,  0,  0}{\textbf{0.919}} & 0.701 & 0.532 & 0.714 & 0.760 & \textcolor[rgb]{ 1,  0,  0}{\textbf{0.932}} & 0.564 & 0.538 & 0.623 & 0.868 \\
    \midrule
    Precision & \textcolor[rgb]{ 1,  0,  0}{\textbf{0.898}} & 0.639 & 0.531 & 0.660 & 0.806 & \textcolor[rgb]{ 1,  0,  0}{\textbf{0.935}} & 0.541 & 0.668 & 0.577 & 0.904 \\
    \midrule
    Recall & \textcolor[rgb]{ 1,  0,  0}{\textbf{0.947}} & 0.925 & 0.600 & 0.884 & 0.683 & \textcolor[rgb]{ 1,  0,  0}{\textbf{0.959}} & 0.858 & 0.238 & 0.917 & 0.823\\
    \midrule
    F1-score & \textcolor[rgb]{ 1,  0,  0}{\textbf{0.922}} & 0.755 & 0.537 & 0.755 & 0.739 & \textcolor[rgb]{ 1,  0,  0}{\textbf{0.933}} & 0.663 & 0.329 & 0.708 & 0.861 \\
    \bottomrule
    \bottomrule
    \end{tabular}%
  \label{tab:5}%
\end{table*}%

\textbf{Author name disambiguation}. To gain the complete and accurate publication records for each scholar, we need to infer their actual identities. Since the MAG dataset does not contain unique author identifiers, we conduct name disambiguation to overcome the problem of duplicate names. Accordingly to Sinatra et al \cite{sinatra2016quantifying}, there are two main steps in the author name disambiguation: separation and mergence. The first step is separating all authors apart. It means that we should regard each author in the records as a unique one. The ultimate goal is to reduce duplicate identifies by merging authors iteratively. We consider the authors with identical names as the same individual if they meet one of the following criteria:
\begin{enumerate}
  \item The two authors have been cited each other at least once;
  \item The two authors have at least one co-author;
  \item The two authors have at least one identical affiliation.
\end{enumerate}
The name disambiguation process ends up with the condition that there are no author pairs to merge.

The matching process consists of two steps: the first step is to find the scholar in the MAG according to scholars' names obtained from AFT; the second step is to obtain features we need, such as publications and collaborators of these scholars. In the first step, we adopt a regular expression to present the name of each advisee and match them in the MAG dataset in their own fields. For example, if the scholars crawled from FTA major in computer science, then we match them in the field of computer science in the MAG dataset. We add all matching results in a dictionary as ``the scholar's name in AFT: the scholar's name in the MAG", where ``the scholar's name in the AFT" is the primary key of the dictionary. In addition, we establish a set to store all the information of the matched advisees. We match the advisor's name from matched advisee's collaborators in the same way. If we can match the advisor in the advisee's collaborators, we define these two collaborators as the ground truth advisor-advisee pair. We use this method to ensure that the matching of advisor-advisee pairs in the MAG dataset is what we really need. Finally, we can obtain the relevant information and digitalize it as the features for training.

\textbf{Disciplinary differences elimination}. As shown in TABLE~\ref{tab:4}, the training dataset contains six independent research fields. To observe the model performance, experiments are conducted independently in each field. When applying the model to the entire dataset, we add a disciplinary parameter $\delta^{f}_{y}$ for the field $f$ in $y$ ($y$ represents the year), which is calculated as:
\begin{equation}
\label{eq15}
\delta^{f}_{y} = |F| * \frac{np^{f}_{y}}{\langle np\rangle_{y}}
\end{equation}
where $np^{f}_{y}$ is the total number of publications in $y$ for $f$, and $\langle np\rangle_{y}$ is the average $np$ calculated over all fields published in $y$. $|F|$ is the number of research categories. Since the papers in the MAG dataset are divided into 19 major disciplines, thus $|F| = 19$ in this paper.

\textbf{Re-scaled number of publications}. From Fig.~\ref{fig:4}, we observe that the growth of publications is in line with exponential distribution. The equation achieved by curve fitting is:
\begin{equation}
\label{eq16}
p = 8.3 * 10^{-45}e^{0.05y} + 22.7*10^{3}
\end{equation}
where $p$ is the number of publication in year $y$. In order to compare the collaboration attributes between advisors and advisees starting their collaboration at the different time period, we use a re-scaled parameter, $\tilde{p}$, to gauge the publication records for each scholar:
\begin{equation}
\label{eq17}
\tilde{p} = p / \tau
\end{equation}
where $\tau$ is the temporal correction factor, which is calculated as:
\begin{equation}
\label{eq18}
\tau = y' = 4.15*10^{-46}e^{0.05y}.
\end{equation}
When calculating the collaboration features, we use the re-scaled dataset to prevent the model performance from temporal bias.

\begin{figure}[!ht]
  \centering
  \includegraphics[width=0.5\textwidth]{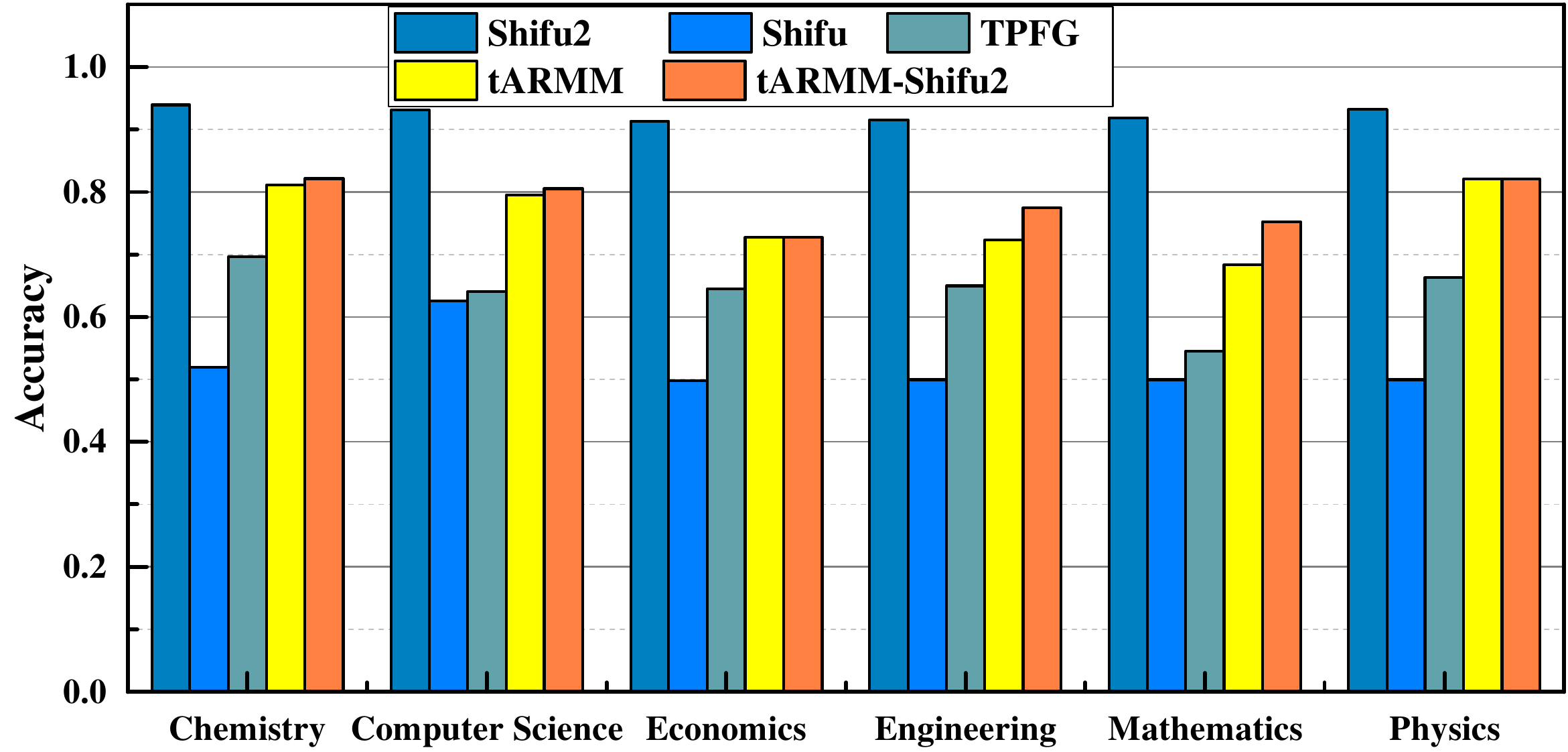}\\
  \caption{Performance in terms of accuracy.}
  \label{fig:11}
\end{figure}

\subsubsection{\textbf{Baselines}}
\label{secb2}
We compare Shifu2 with the following baseline models:
\begin{itemize}
  \item \textbf{DeepWalk}~\cite{perozzi2014deepwalk}. DeepWalk uses random walks to obtain the local information of the network and treats the walks as equivalent sentences. By employing skip-gram, it learns the latent representations for vertices.
  \item \textbf{LINE}~\cite{tang2015line}. LINE eliminates the limitation of classical stochastic gradient descent by using the edge-sampling algorithm. It can preserve both local and global structures for following networks: direct/undirect networks and weighted/unweighted networks.
  \item \textbf{Node2vec}~\cite{grover2016node2vec}. Node2vec is a semi-supervised method focusing on preserving neighbors' information for each node in the network. It can capture the diversity of connectivity patterns and learn low-dimensional feature representations for nodes effectively.
  \item \textbf{TransNet}~\cite{tu2017transnet}. TransNet is a knowledge graph-based framework, which translates the interactions between vertices to a translation operation. It considers the semantic information for each edge as a binary.
\end{itemize}
We also compare Shifu2 with existing solutions for identifying advisor-advisee relationships:
\begin{itemize}
  \item \textbf{Shifu}~\cite{wang2017shifu}. Shifu is a deep learning model based on the stacked autoencoder. It takes scholars' personal properties and network characteristics as the input.
  \item \textbf{TPFG}~\cite{wang2010mining}. TPFG is a time-constrained probabilistic model. It considers the task of advisor-advisee relationship as a jointly likelihood objective optimization problem.
  \item \textbf{tARMM}~\cite{zhao2018identifying}. tARMM is a deep model based on the improved Refresh Gate Recurrent Units. It is inspired by the idea of variance Recurrent Neural Network models.
  \item \textbf{tARMM-Shifu2}. In tARMM-Shifu2, we modify the input of tARMM. We use node and edge attributes proposed in Shifu2 as the input of tARMM.
\end{itemize}
\begin{table}[htbp]
  \centering
  \caption{Number of hidden units in each layer for node autoencoder and edge autoencoder}
    \begin{tabular}{c|c|c|l}
    \toprule
    \multicolumn{1}{c|}{Encoder types}&Group&No. layers&\multicolumn{1}{c}{No. units}\\
    \midrule
    \midrule
    \multirow{5}[10]{*}{Node encoder} &1    &1         &2000                         \\
\cmidrule{2-4}                        &2    &2         &2000, 1000                   \\
\cmidrule{2-4}                        &3    &3         &2000, 1000, 500              \\
\cmidrule{2-4}                        &4    &4         &2000, 1500, 1000, 500        \\
\cmidrule{2-4}                        &5    &5         &2000, 1500, 1000, 500, 300   \\
    \midrule
    \multirow{5}[10]{*}{Edge encoder} &1    &1         &18                           \\
\cmidrule{2-4}                        &2    &2         &18, 50                       \\
\cmidrule{2-4}                        &3    &3         &18, 50, 70                   \\
\cmidrule{2-4}                        &4    &4         &18, 30, 50, 70               \\
\cmidrule{2-4}                        &5    &5         &18, 30, 50, 70, 90           \\
    \bottomrule
    \bottomrule
    \end{tabular}%
  \label{tab:6}%
\end{table}%

\begin{figure*}[htbp]
\centering
\subfigure[Accuracy]{
\label{fig:6-a}
\includegraphics[width=0.32\textwidth]{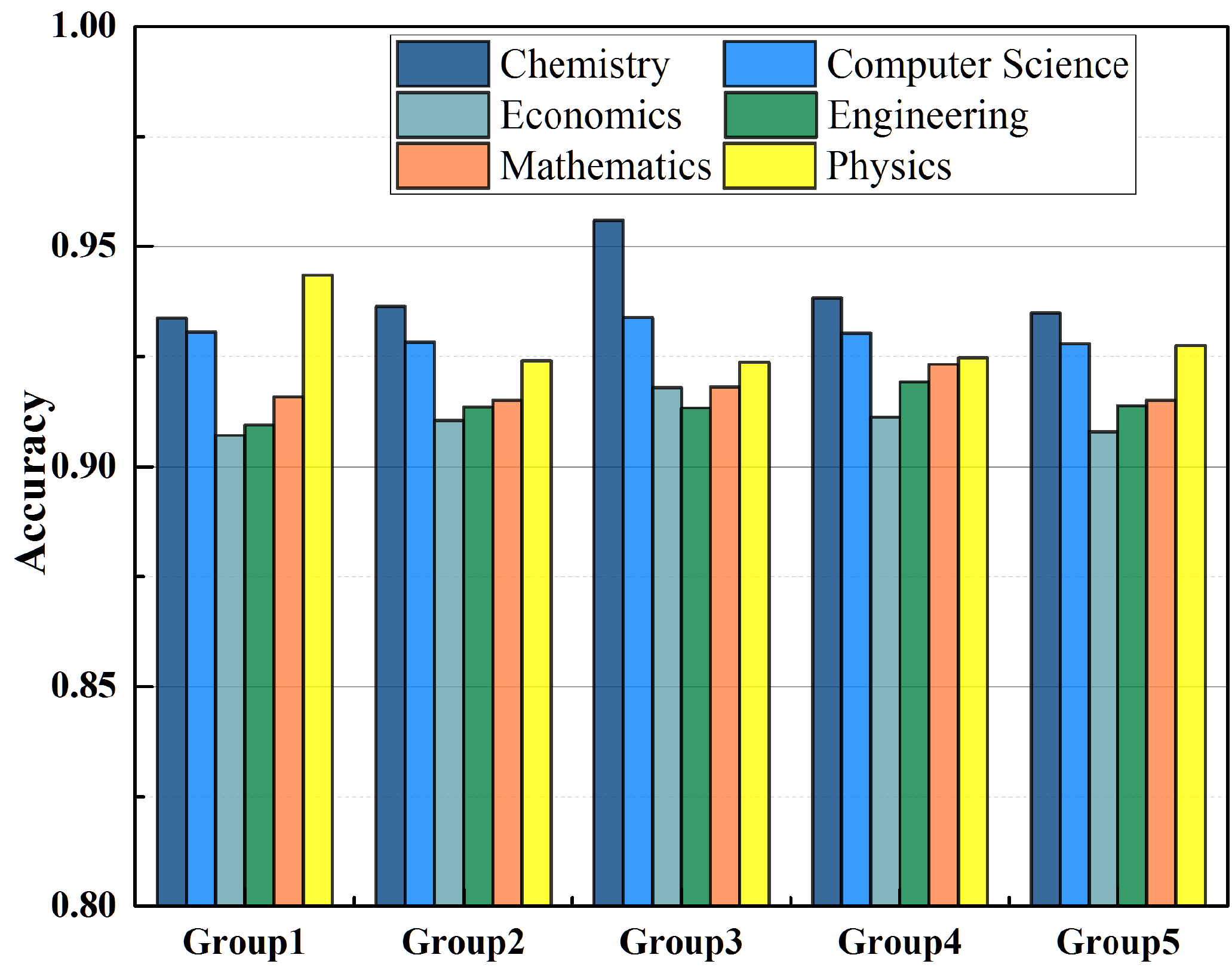}}
\subfigure[Recall]{
\label{fig:6-b}
\includegraphics[width=0.32\textwidth]{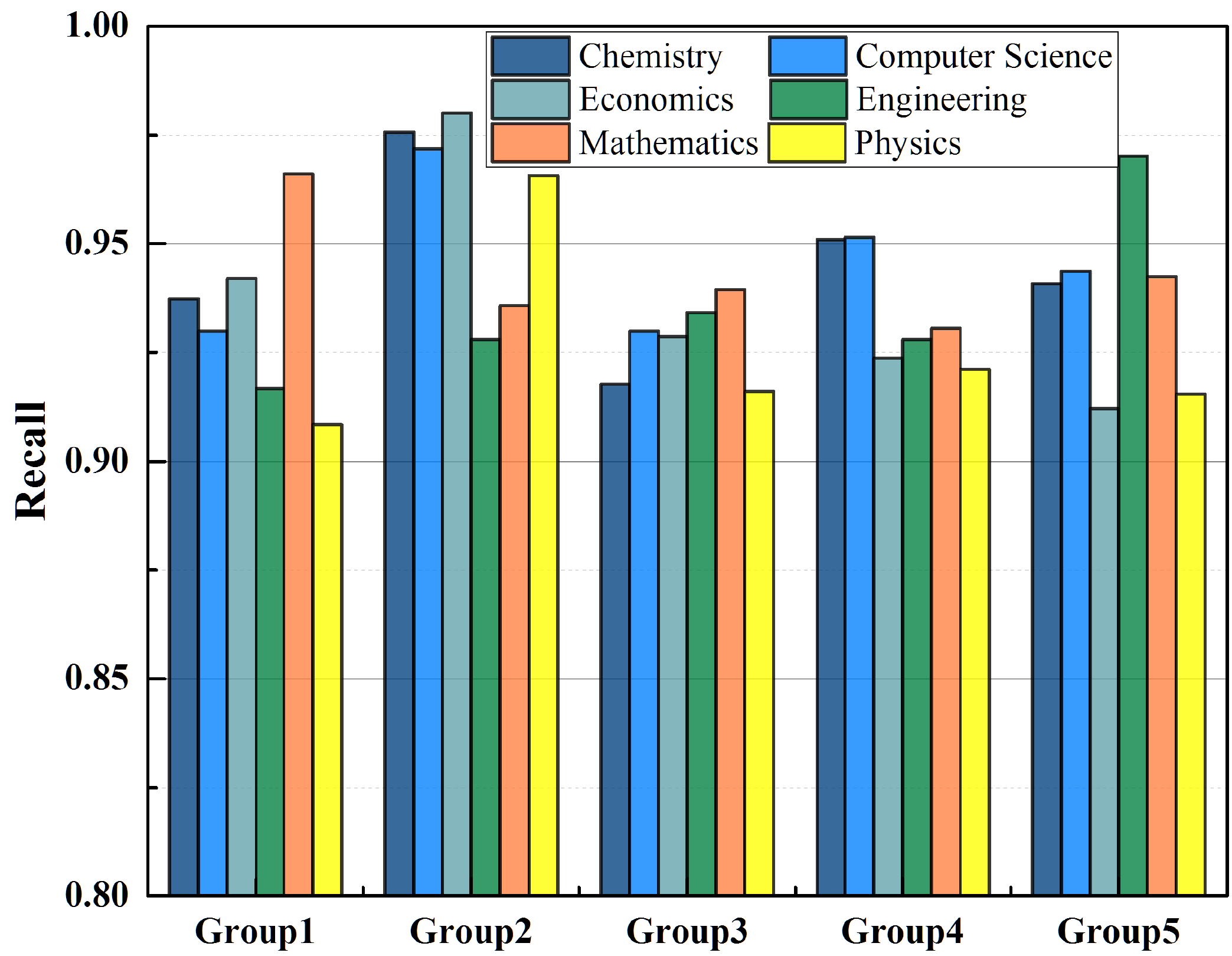}}
\subfigure[F1-score]{
\label{fig:6-c}
\includegraphics[width=0.32\textwidth]{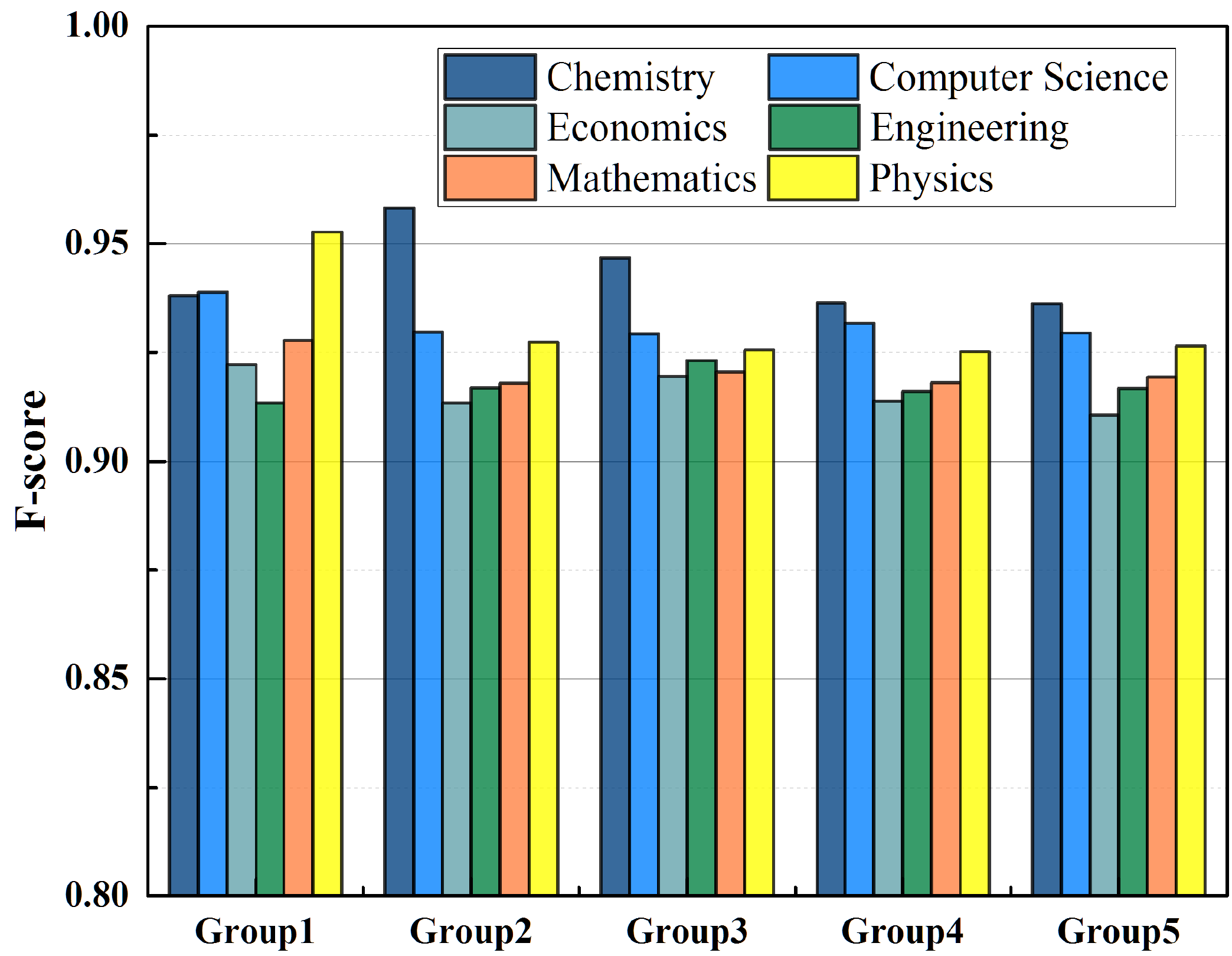}}
\caption{Shifu2 performance with different hidden layers in node autoencoder. }
\label{fig:6}
\end{figure*}

\subsubsection{\textbf{Evaluation Metrics and Parameter Settings}}
Since the task of advisor-advisee relationship mining can be regarded as a binary prediction problem, i.e., for each collaboration pair $(i,j)$, this is the binary classification of whether $j$ is $i$'s advisor. To evaluate the performance of Shifu2, four widely used metrics for classification tasks are used in our experiments: Accuracy, Precision, Recall, and F1-score.
\begin{figure*}[!ht]
\centering
\subfigure[Accuracy]{
\label{fig:7-a}
\includegraphics[width=0.32\textwidth]{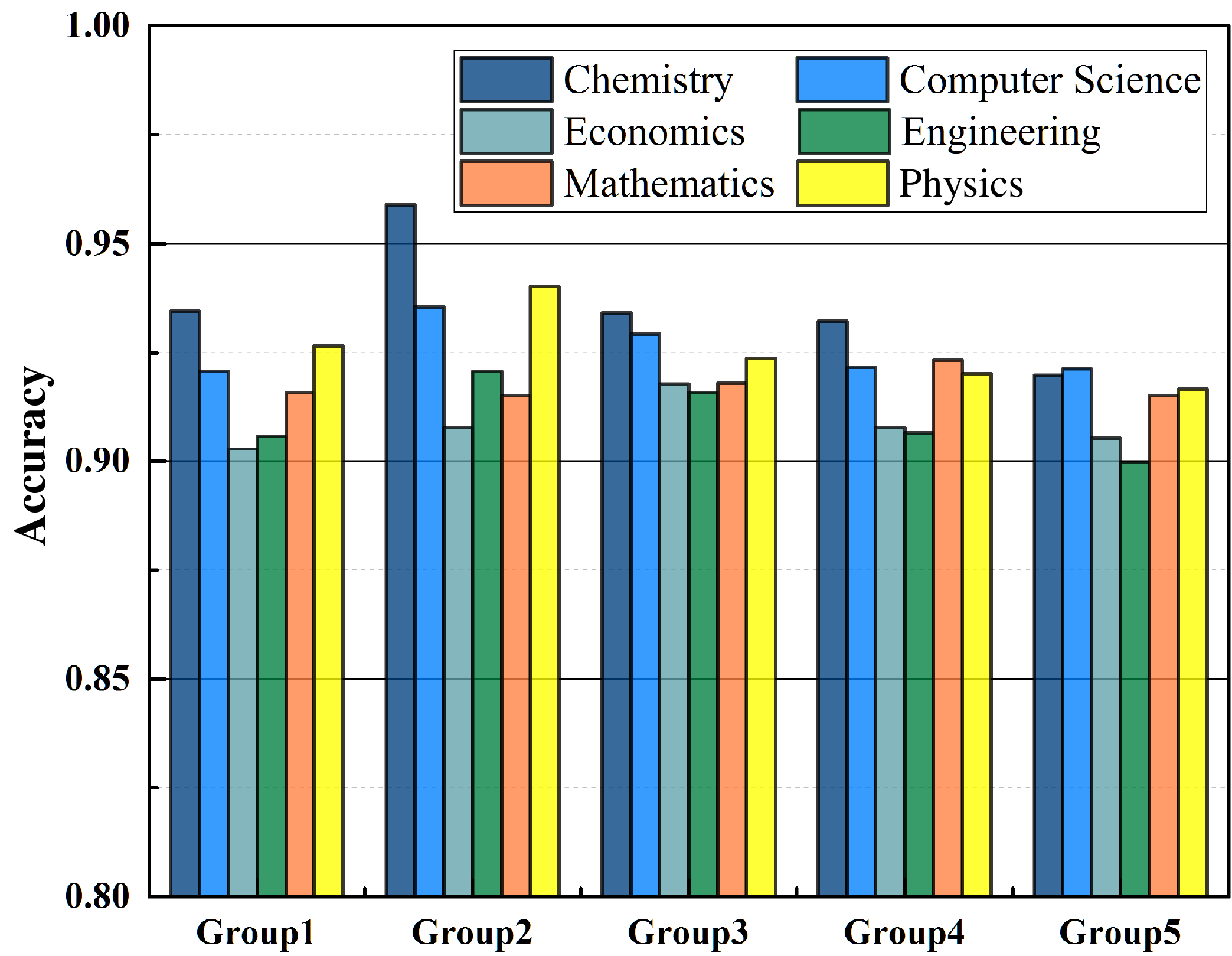}}
\subfigure[Recall]{
\label{fig:7-b}
\includegraphics[width=0.32\textwidth]{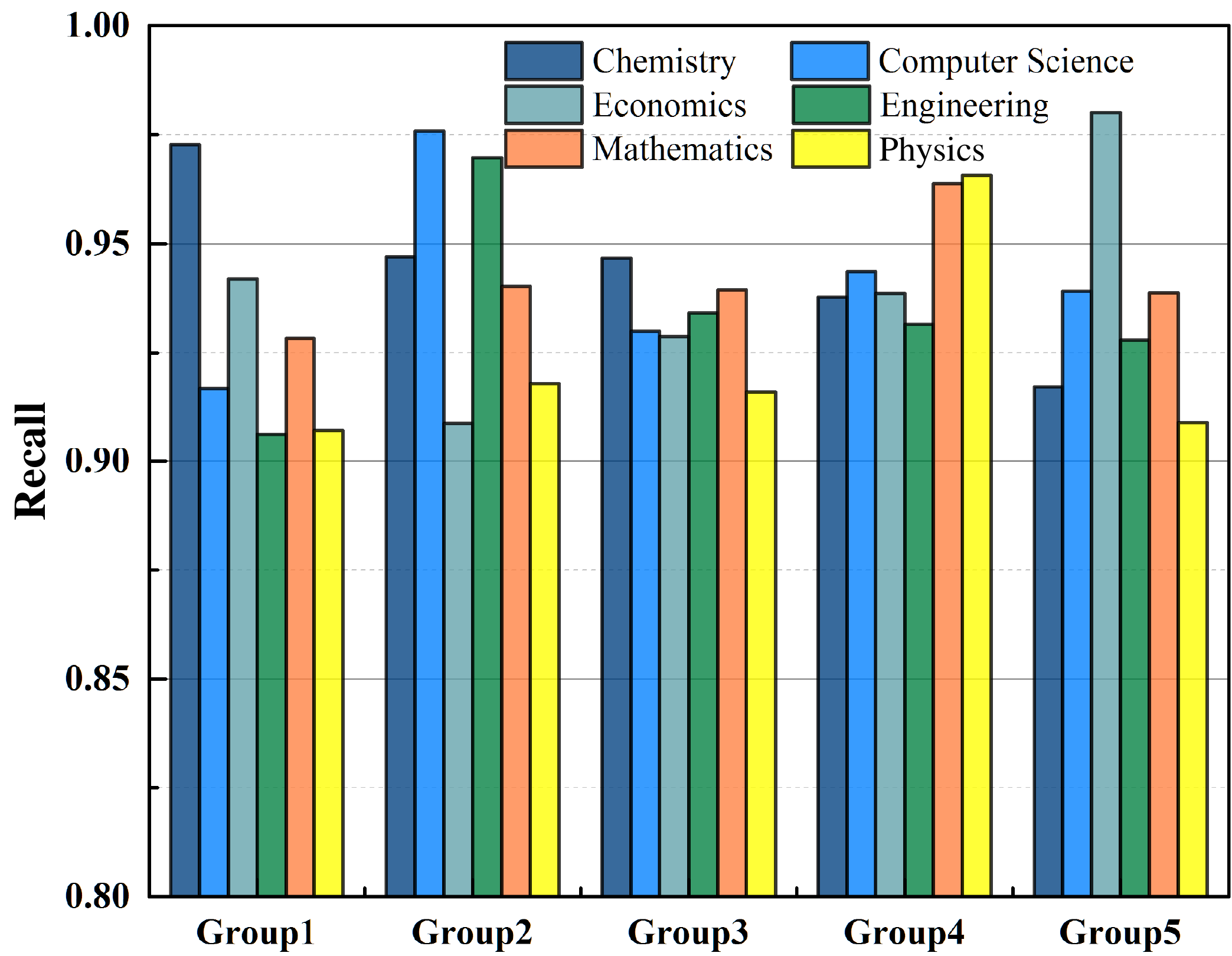}}
\subfigure[F1-score]{
\label{fig:7-c}
\includegraphics[width=0.32\textwidth]{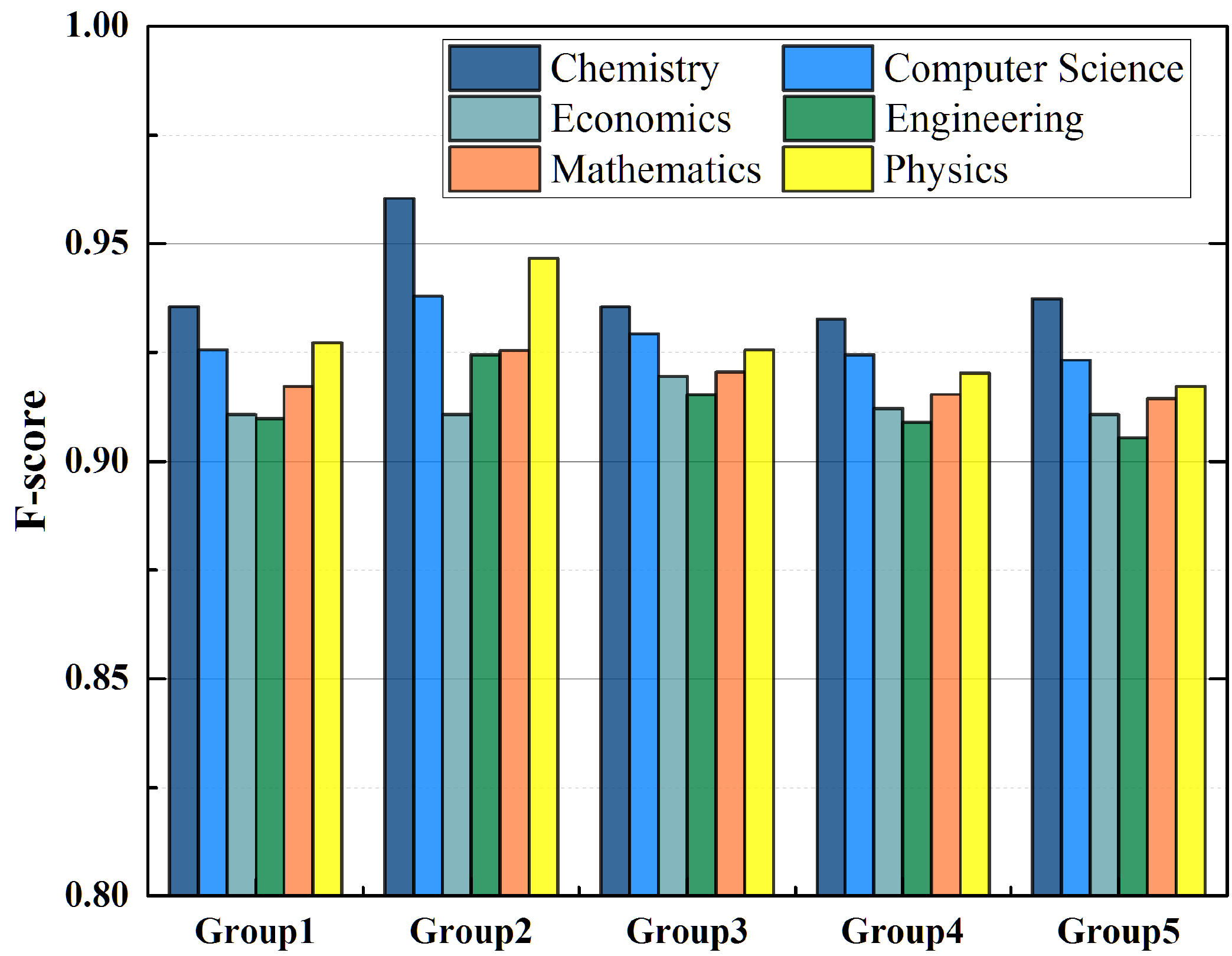}}
\caption{Shifu2 performance with different hidden layers in edge autoencoder.}
\label{fig:7}
\end{figure*}

\begin{figure*}[htbp]
\centering
\subfigure[Accuracy]{
\label{fig:8-a}
\includegraphics[width=0.32\textwidth]{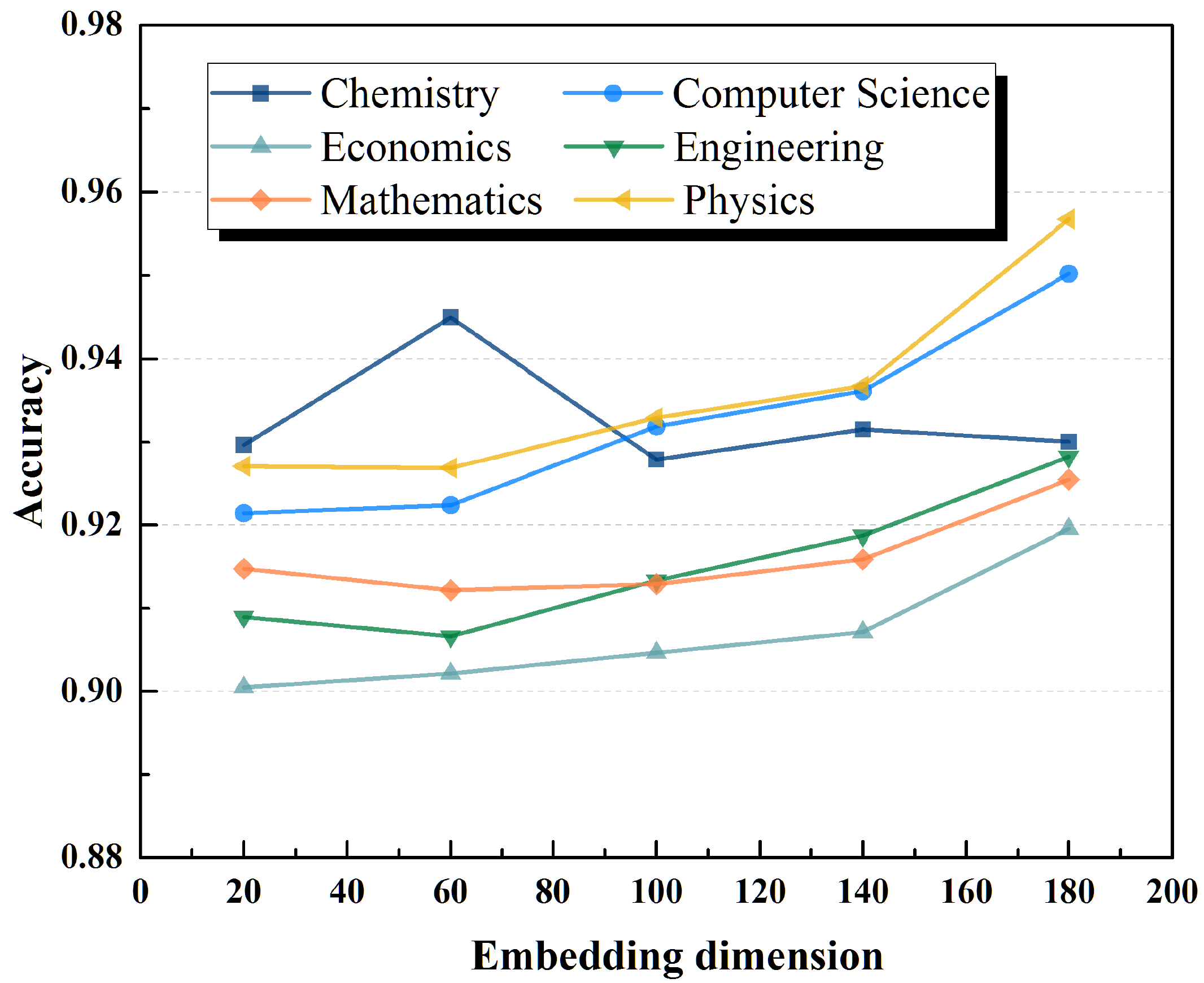}}
\subfigure[Recall]{
\label{fig:8-b}
\includegraphics[width=0.32\textwidth]{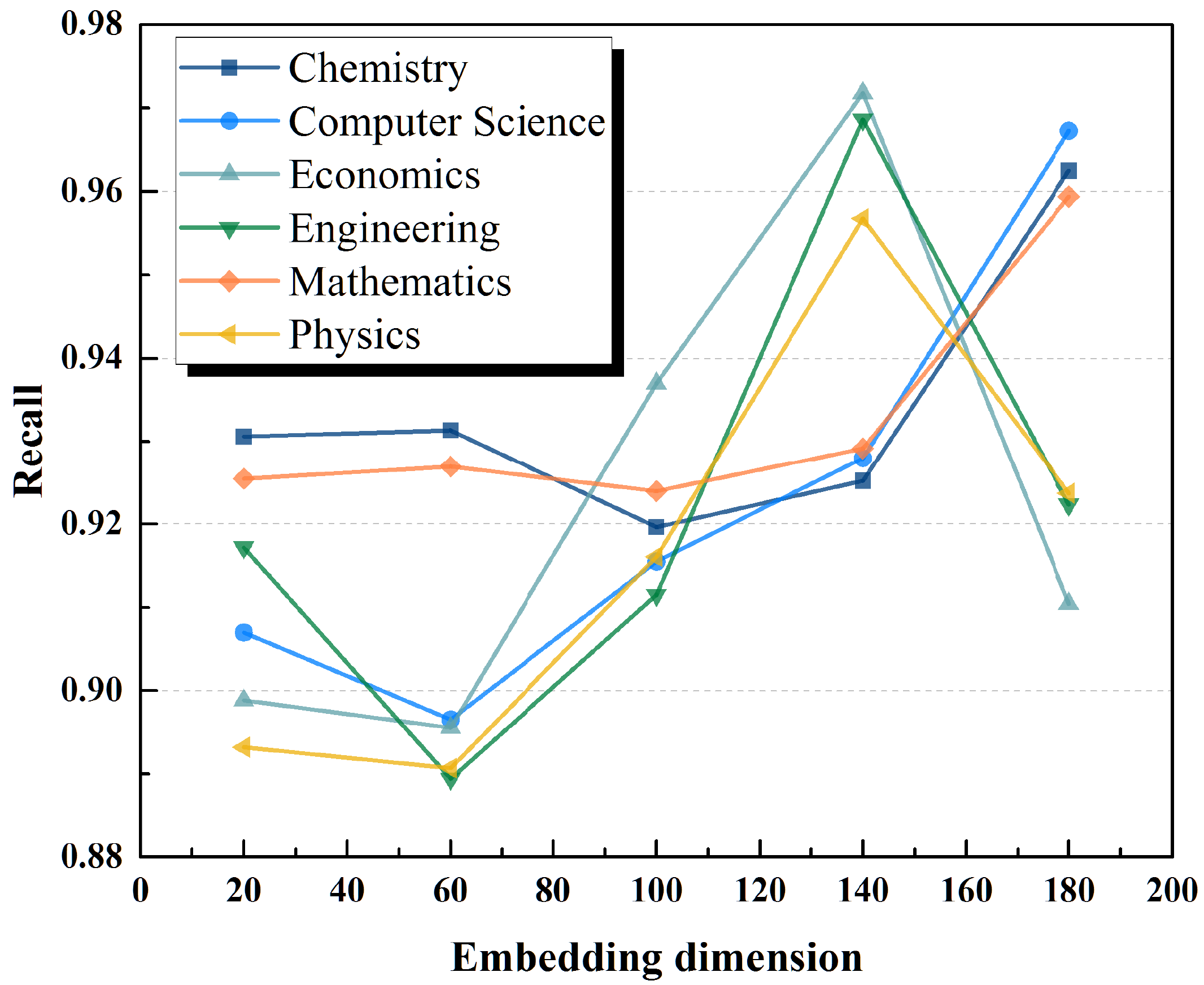}}
\subfigure[F1-score]{
\label{fig:8-a}
\includegraphics[width=0.32\textwidth]{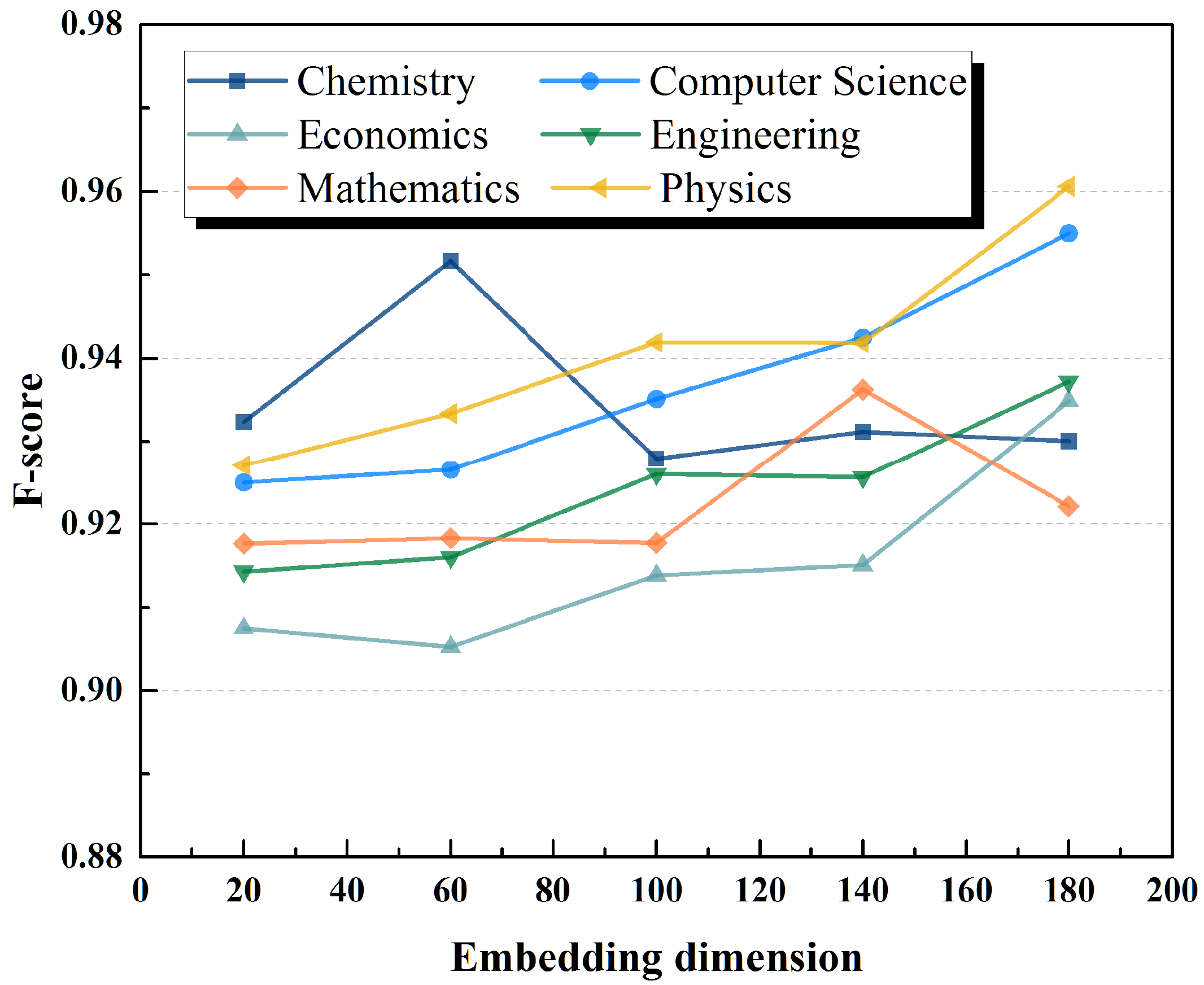}}
\caption{Shifu2 performance with respect to embedding dimension in node autoencoder and edge autoencoder. }
\label{fig:8}
\end{figure*}

\subsection{Results and Analysis}
To evaluate the effectiveness of Shifu2 on identifying the advisor-advisee relationships, we conduct a series of experiments from the following aspects. First, in order to illustrate how much Shifu2 can improve the embedding representation, we compare our proposed model with state-of-the-art NRL models mentioned in Section~\ref{secb2}. Then we try to explore the effect of different assumptions in affecting the effectiveness of the model. Meanwhile, we compare the model performance between different research fields to find out the disciplinary differences.
\begin{table*}[htb]
  \centering
  \caption{Performance of Shifu2 with different learning rates}
    \begin{tabular}{c|c|c|c|c|c|c|c|c|c|c|c|c}
    \toprule
          & \multicolumn{4}{c|}{Chemistry} & \multicolumn{4}{c|}{Computer Science} & \multicolumn{4}{c}{Economics} \\
    \midrule
     \diagbox{Metrics}{Learning rate}  & 0.001  & 0.005  & 0.01  & 0.1   & 0.001  & 0.005  & \multicolumn{1}{l|}{0.01 } & 0.1   & 0.001  & 0.005  & 0.01  & 0.1  \\
    \midrule
    Accuracy & 0.936 & 0.943 & \textbf{0.942} & 0.932 & 0.911 & 0.921 & \textbf{0.931} & 0.929 & 0.892 & 0.905 & \textbf{0.911} & 0.911 \\
    \midrule
    Precision & 0.943 & 0.945 & 0.937 & 0.919 & 0.928 & 0.929 & 0.926 & 0.905 & 0.880 & 0.888 & 0.886 & 0.880 \\
    \midrule
    Recall & 0.932 & 0.945 & 0.949 & 0.948 & 0.896 & 0.916 & 0.941 & 0.958 & 0.920 & 0.934 & 0.949 & 0.9534 \\
    \midrule
    F1-score & 0.938 & 0.945 & 0.943 & 0.933 & 0.911 & 0.923 & 0.933 & 0.931 & 0.898 & 0.910 & 0.916 & 0.915 \\
    \midrule
          & \multicolumn{4}{c|}{Engineering} & \multicolumn{4}{c|}{Mathematics} & \multicolumn{4}{c}{Physics} \\
    \midrule
     \diagbox{Metrics}{Learning rate}      & 0.001  & 0.005  & 0.01  & 0.1   & 0.001  & 0.005  & 0.01  & 0.1   & 0.001  & 0.005  & 0.01  & 0.1  \\
    \midrule
    Accuracy & 0.902 & 0.914 & \textbf{0.916} & 0.912 & 0.911 & 0.914 & 0.917 &\textbf{ 0.919} & 0.915 & 0.937 & \textbf{0.943} & 0.926 \\
    \midrule
    Precision & 0.905 & 0.905 & 0.902 & 0.888 & 0.903 & 0.905 & 0.899 & 0.891 & 0.932 & 0.955 & 0.953 & 0.924 \\
    \midrule
    Recall & 0.905 & 0.928 & 0.937 & 0.944 & 0.925 & 0.928 & 0.942 & 0.956 & 0.900 & 0.921 & 0.933 & 0.928 \\
    \midrule
    F1-score & 0.905 & 0.916 & 0.919 & 0.915 & 0.914 & 0.916 & 0.920 & 0.922 & 0.916 & 0.938 & 0.943 & 0.926 \\
    \bottomrule
    \bottomrule
    \end{tabular}%
  \label{tab:7}%
\end{table*}%

\subsubsection{\textbf{Effectiveness Evaluation}}
TABLE~\ref{tab:5} presents the results of comparing Shifu2 with all NRL based baselines. Note that, in the process of parameters setting in node2vec, $p$ and $q$ are selected in the set $\{0.25, 0.5, 1, 2, 4\}$. Finally, they are decided by the maximum of accuracy for each discipline. From the results, we can see that the task of advisor-advisee relationship identification achieves significant improvements by our proposed model. It demonstrates the effectiveness of Shifu2.

Besides, we vary the nodes for training from 10\% to 100\% of the training set. From Fig.~\ref{fig:5} we can observe that Shifu2 consistently achieves stable performance in contrast to other baselines under different sizes of training data. Taking the accuracy rate as an example, Shifu2 is 30\%, 38\%, 28\%, and 8\% higher than deepwalk, line, node2vec, and Transnet in the field of Chemistry, respectively.

Finally, we compare our model with existing methods for advisor-advisee relationships identification mentioned in Section~\ref{secb2}. As shown in Fig.~\ref{fig:11}, Shifu2 achieves better performance consistently across all fields. Furthermore, we can observe that by using the node attributes and edges attributes proposed in Shifu2, the performance of the existing method such as tARMM has been improved.

\subsubsection{\textbf{Parameter Sensitivity}}
In this subsection, we study the effect of the following parameters: (1) number of hidden layers of the node auto-encoder, (2) number of hidden layers of the edge autoencoder, (3) learning rate, (4) number of nodes used for training, (5) embedding dimension, and (6) input features.

\begin{table*}[htbp]
  \centering
  \caption{Performance of Shifu2 with different sizes of data}
    \begin{tabular}{c|c|c|c|c|c|c|c|c|ccc|c}
    \toprule
          & \multicolumn{4}{c|}{Chemistry} & \multicolumn{4}{c|}{Computer Science} & \multicolumn{4}{c}{Economics} \\
    \midrule
     \diagbox{Metrics}{Percentage}      & 20\%  & 40\%  & 60\%  & 80\%  & 20\%  & 40\%  & 60\%  & 80\%  & \multicolumn{1}{c|}{20\%} & \multicolumn{1}{c|}{40\%} & 60\%  & 80\% \\
    \midrule
    Accuracy & 0.935 & 0.935 & \textbf{0.939} & 0.938 & 0.927 & 0.930 & 0.931 & \textbf{0.931} & \multicolumn{1}{c|}{0.908} & \multicolumn{1}{c|}{0.909} & \textbf{0.913} & 0.912 \\
    \midrule
    Precision & 0.919 & 0.919 & \textbf{0.925} & 0.923 & 0.903 & 0.912 & 0.908 & \textbf{0.912} & \multicolumn{1}{c|}{0.875} & \multicolumn{1}{c|}{0.882} & 0.886 & \textbf{0.877} \\
    \midrule
    Recall & 0.955 & 0.955 & \textbf{0.958} & 0.957 & 0.959 & 0.953 & \textbf{0.959} & 0.953 & \multicolumn{1}{c|}{0.955} & \multicolumn{1}{c|}{0.947} & 0.951 & \textbf{0.961} \\
    \midrule
    F1-score & 0.937 & 0.936 & \textbf{0.941} & 0.940 & 0.930 & 0.932 & \textbf{0.933} & 0.932 & \multicolumn{1}{c|}{0.913} & \multicolumn{1}{c|}{0.913} & \textbf{0.917} & 0.917 \\
    \midrule
          & \multicolumn{4}{c|}{Engineering} & \multicolumn{4}{c|}{Mathematics} & \multicolumn{4}{c}{Physics} \\
    \midrule
    \diagbox{Metrics}{Percentage}   & 20\%  & 40\%  & 60\%  & 80\%  & 20\%  & 40\%  & 60\%  & 80\%  & \multicolumn{1}{c|}{20\%} & \multicolumn{1}{c|}{40\%} & 60\%  & 80\% \\
    \midrule
    Accuracy & 0.911 & 0.913 & \textbf{0.915} & 0.915 & 0.916 & 0.918 & 0.919 & \textbf{0.919} & \multicolumn{1}{c|}{0.927} & \multicolumn{1}{c|}{0.929} & \textbf{0.932} & 0.930 \\
    \midrule
    Precision & 0.892 & 0.889 & \textbf{0.889} & 0.888 & 0.892 & 0.895 & 0.897 & \textbf{0.898} & \multicolumn{1}{c|}{0.925} & \multicolumn{1}{c|}{0.931} & 0.928 & \textbf{0.935} \\
    \midrule
    Recall & 0.938 & 0.945 & 0.950 & \textbf{0.952} & 0.948 & \textbf{0.950 }& 0.947 & 0.947 & \multicolumn{1}{c|}{0.931} & \multicolumn{1}{c|}{0.928} & \textbf{0.937} & 0.925 \\
    \midrule
    F1-score & 0.914 & 0.916 & \textbf{0.919} & 0.919 & 0.919 & 0.921 & 0.921 & \textbf{0.922} & 0.928 & 0.929 & \textbf{0.933} & 0.930 \\
    \bottomrule
    \bottomrule
    \end{tabular}%
  \label{tab:8}%
\end{table*}%

\textbf{Number of hidden layers}. While the training set is determined, we should consider how to grid-search the number of hidden layers. Studies demonstrate that models will achieve better performance with more hidden layers. The highest accuracy can even reach 100\%. However, models will be prone to over-fitting, thus the prediction performance will be sharply reduced on the test data. So we need to consider the effective depth of the training to make Shifu2 achieve better prediction performance without over-fitting. The number of units in each hidden layer needs to be determined accordingly.

In this paper, we need to determine the number of hidden layers for the node autoncoder and edge autoencoder, respectively. For the node autoencoder and the edge autoencoder, we choose the hidden layers from 1 to 5. The number of hidden units in each layer is presented in TABLE~\ref{tab:6}. Fig.~\ref{fig:6} and Fig.~\ref{fig:7} present the experimental results. Thus we can obtain the best architecture of our proposed model with three hidden layers for the node autoencoder, and two layers for the edge autoencoder.

\textbf{Learning rate}. Learning rate is a crucial parameter in Shifu2 because it controls the update speed of the model. If the learning rate is overlarge, the value of the loss function will move backwards and forwards around the minimum and will not converge. Otherwise, it will lead to a slow learning process. In TABLE~\ref{tab:7}, we discover that if the learning rate is set to 0.01, the model can achieve the best performance across almost all fields.
\begin{table}[htbp]
  \centering
  \caption{Performance of Shifu2 without node autoencoder}
    \begin{tabular}{c|c|c|c|c}
    \toprule
          & \multicolumn{2}{c|}{Chemistry} & \multicolumn{2}{c}{Computer Science} \\
    \midrule
      \diagbox{Metrics}{Percentage} & Shifu2 & Shifu2-E & Shifu2 & Shifu2-E \\
    \midrule
    Accuracy & \textbf{0.939} & 0.789 & \textbf{0.931} & 0.813 \\
    \midrule
    Precision & \textbf{0.925} & 0.753 & \textbf{0.912} & 0.782 \\
    \midrule
    Recall & \textbf{0.958} & 0.914 & \textbf{0.959} & 0.883 \\
    \midrule
    F1-score & \textbf{0.941} & 0.823 & \textbf{0.933} & 0.830 \\
    \midrule
          & \multicolumn{2}{c|}{Economics} & \multicolumn{2}{c}{Engineering} \\
    \midrule
      \diagbox{Metrics}{Percentage}& Shifu2 & Shifu2-E & Shifu2 & Shifu2-E \\
    \midrule
    Accuracy & \textbf{0.913} & 0.507 & \textbf{0.915} & 0.736 \\
    \midrule
    Precision & \textbf{0.877} & 0.506 & \textbf{0.889} & 0.718 \\
    \midrule
    Recall & \textbf{0.961} & 0.602 & \textbf{0.952} & 0.873 \\
    \midrule
    F1-score & \textbf{0.917} & 0.550 & \textbf{0.919} & 0.784 \\
    \midrule
          & \multicolumn{2}{c|}{Mathematics} & \multicolumn{2}{c}{Physics} \\
    \midrule
      \diagbox{Metrics}{Percentage} & Shifu2 & Shifu2-E & Shifu2 & Shifu2-E \\
    \midrule
    Accuracy & \textbf{0.919} & 0.702 & \textbf{0.932} & 0.846 \\
    \midrule
    Precision & \textbf{0.898} & 0.670 & \textbf{0.935} & 0.827 \\
    \midrule
    Recall & \textbf{0.947} & 0.889 & \textbf{0.959} & 0.890 \\
    \midrule
    F1-score & \textbf{0.922} & 0.760 & \textbf{0.933} & 0.857 \\
    \bottomrule
    \bottomrule
    \end{tabular}%
  \label{tab:9}%
\end{table}%

\textbf{Number of nodes used for training}. We set the number of nodes for training to be 20\%, 40\%, 60\%, and 80\% of the entire training data, respectively, to evaluate the sensitivity with respect to the size of training set. TABLE~\ref{tab:8} presents the identification performance of the model. From the results, we observe that our model is well performed even though the training data is sparse. Shifu2 can reach the accuracy higher than 90\% in all research fields with only 1000 nodes, which indicates the robustness of Shifu2.

\textbf{Embedding dimension}. After learning, we will get node representations and edge representations in terms of $n$-dimensional vectors, where $n$ is the artificially set embedding dimension. To verify whether it influences the performance of the model, we vary it from 20 to 160 in Shifu2. The results are shown in Fig.~\ref{fig:8}. From the results we can see that, our model performs well with both low-dimensional representations and high-dimensional representations, which can be proved by the accuracy higher than 90\% with different embedding dimensions. We also observe that the disciplinary differences do exist. For example, the performance of the model in terms of the accuracy keeps increasing with larger embedding dimensions for all research fields except chemistry. For chemistry, the accuracy first presents an increasing trend. After reaching a peak, it begins to decline.

\textbf{Input features}. A key ingredient of Shifu2 is the exploitation of node attributes and edge attributes. To validate the effectiveness of this mechanism, here we focus on examining the performance of Shifu2 without the node autoencoder (represented as Shifu2-E in TABLE~\ref{tab:9}) first. From TABLE~\ref{tab:9}, we can conclude that if we only use the edge autoencoder, it will achieve a relatively poor performance. The use of node attributes clearly improves the performances.

An advisee is usually supervised by an advisor for a specific period of time (i.e., not forever), during which they collaborate with each other closely. In many (if not most) cases, for instance, it takes about 3 to 5 years for a PhD student to graduate from the university. After graduation, many students carry out their own research work, with much less or even no collaboration with their (PhD) supervisors. As a result, the attributes of collaboration networks will change. For example, collaboration frequency might decrease. On the other hand, it is difficult to determine when advisees will collaborate with their advisors since the collaboration pattern may vary from case to case.

\begin{figure}[htbp]
\centering
\subfigure[Accuracy]{
\label{fig:9-a}
\includegraphics[width=0.23\textwidth]{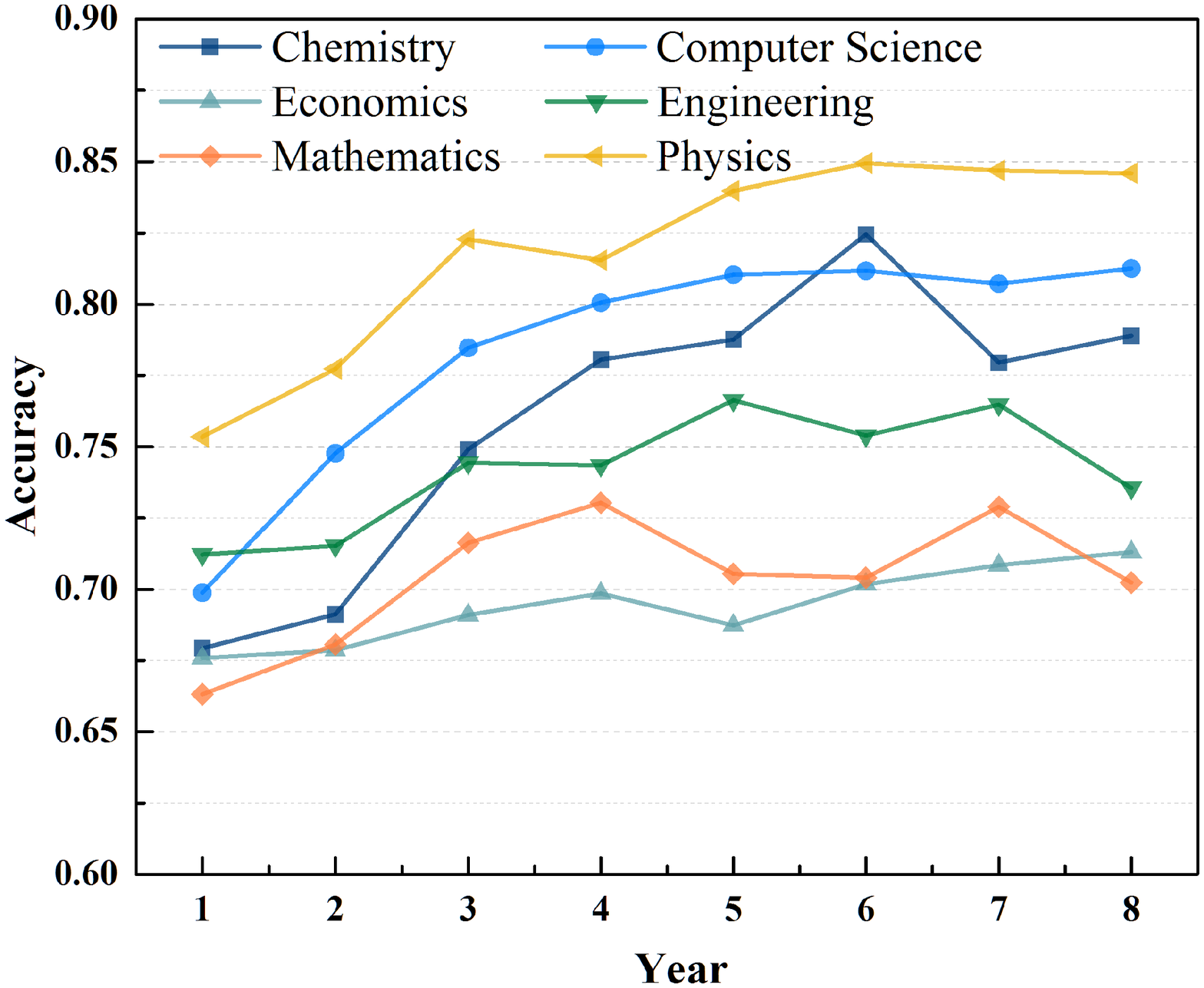}}
\subfigure[Precision]{
\label{fig:9-b}
\includegraphics[width=0.23\textwidth]{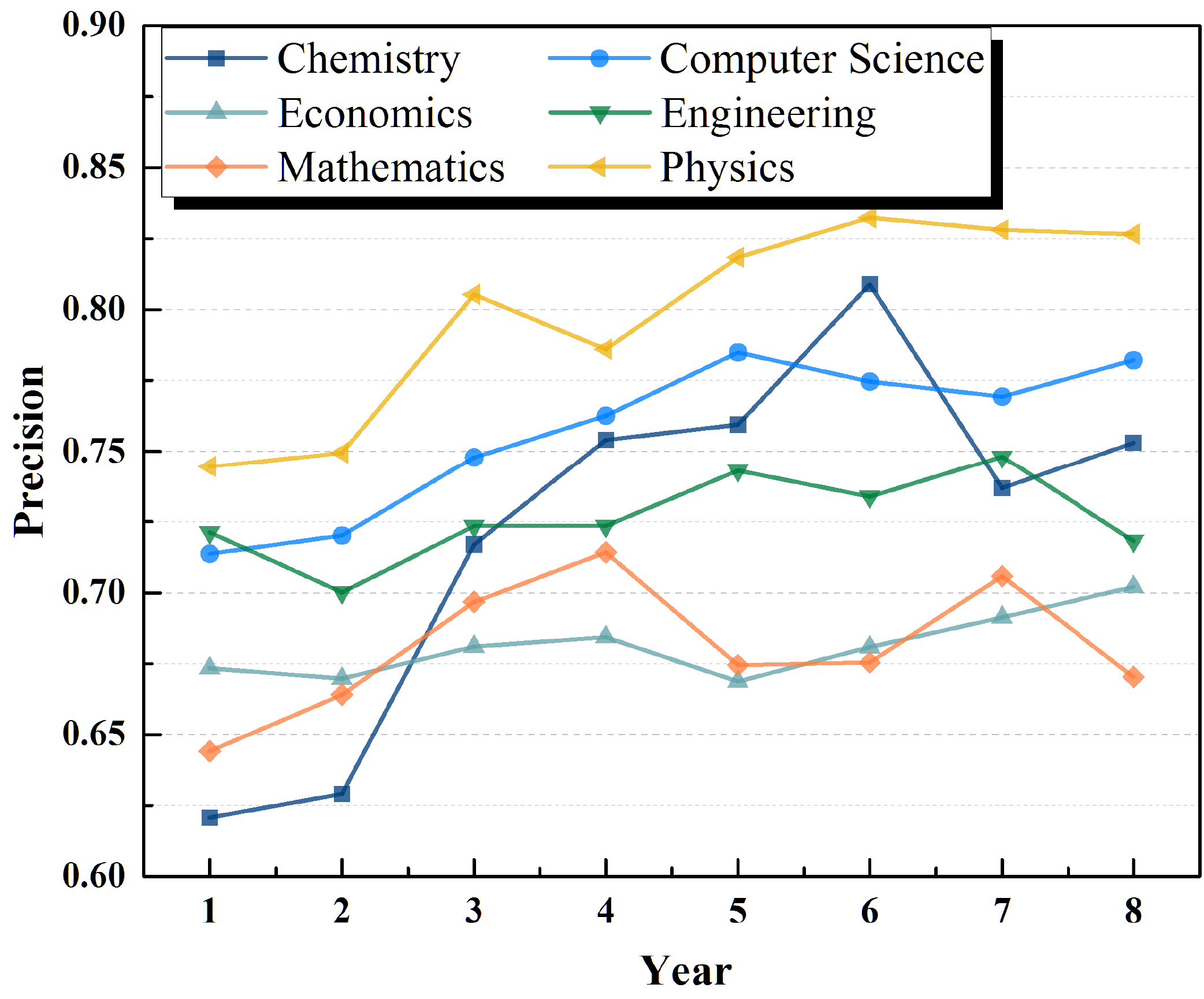}}\\
\subfigure[Recall]{
\label{fig:9-c}
\includegraphics[width=0.23\textwidth]{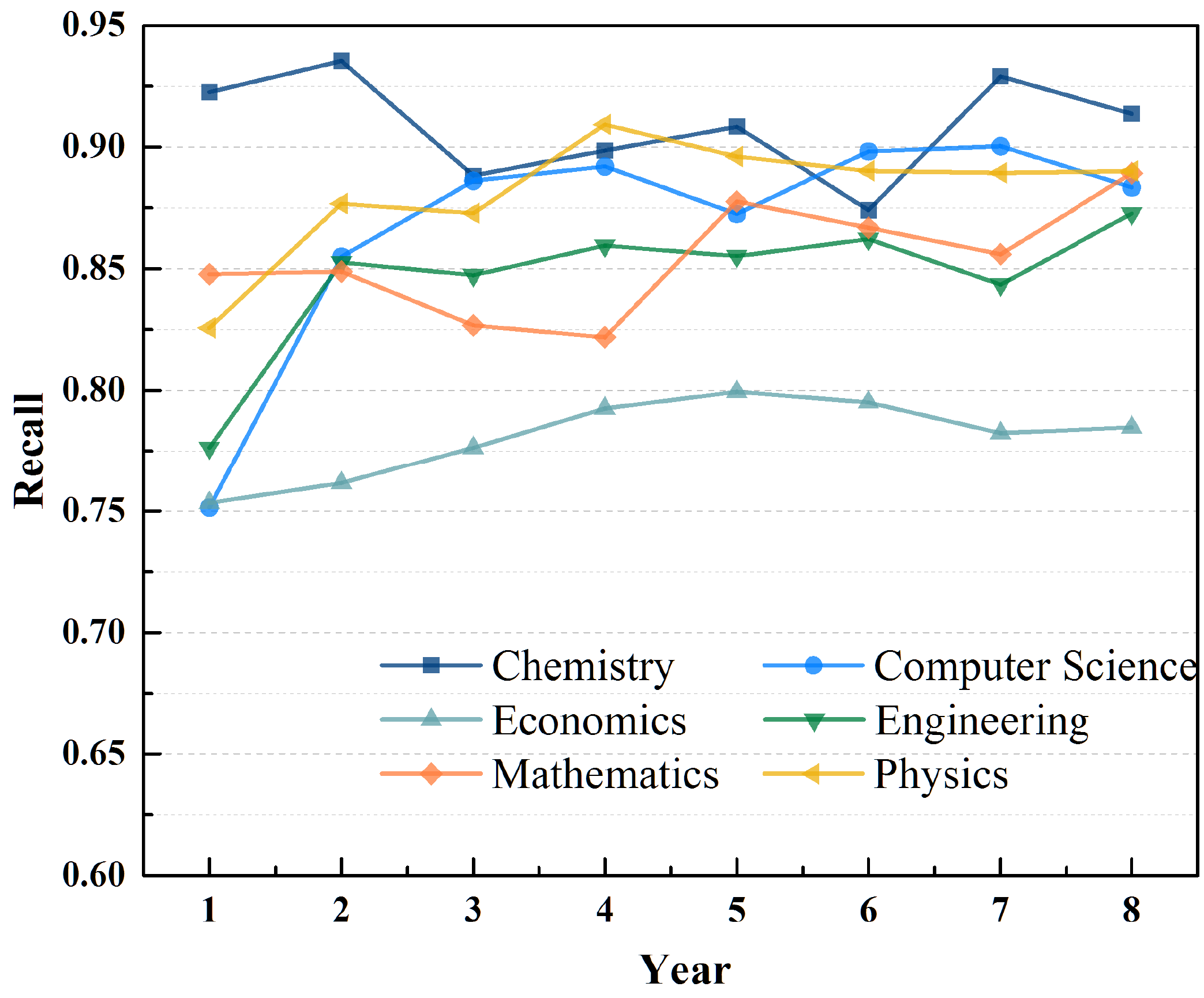}}
\subfigure[F1-score]{
\label{fig:9-d}
\includegraphics[width=0.23\textwidth]{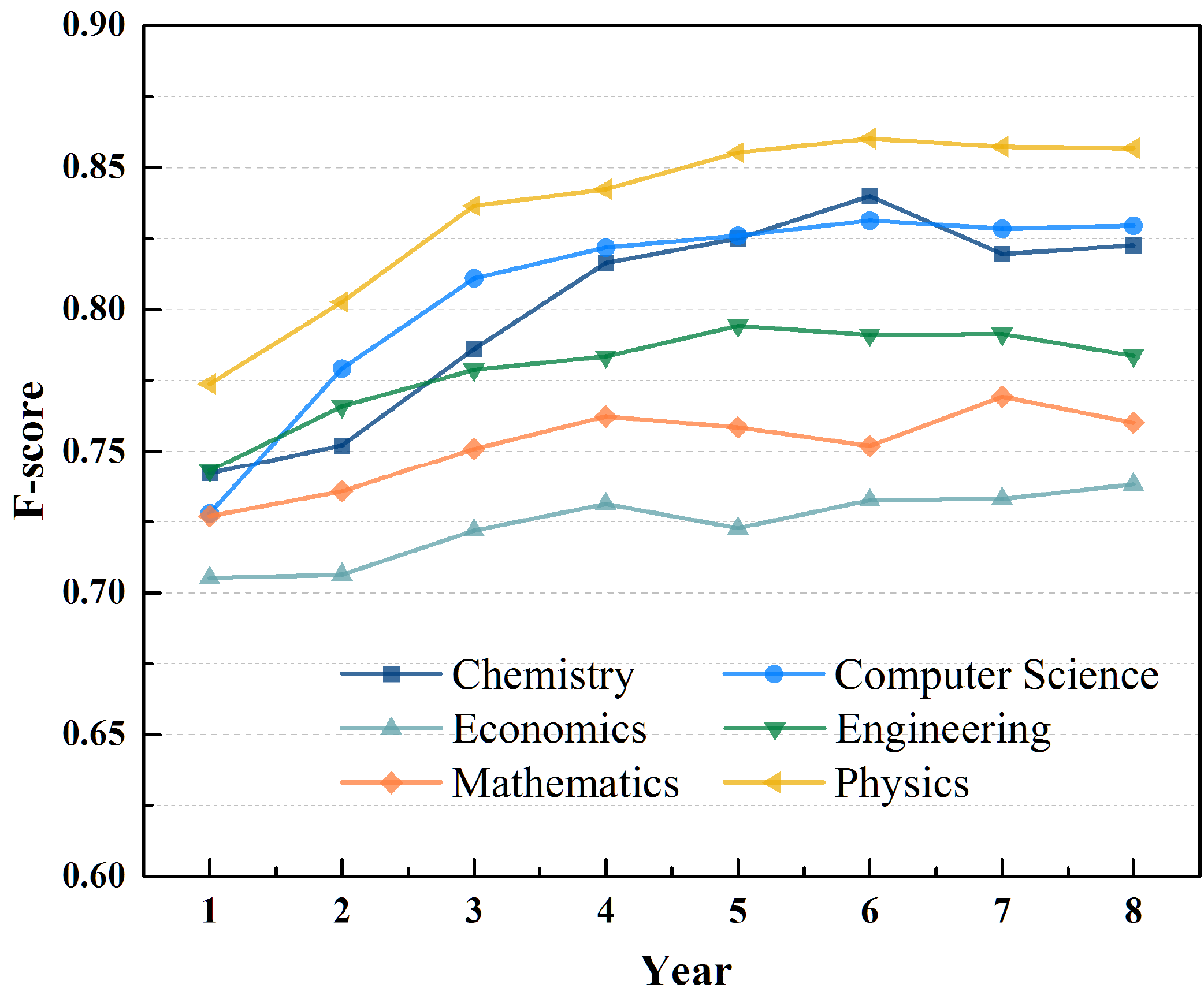}}
\caption{Shifu2 performance with different time lengths of input data in edge autoencoder. }
\label{fig:9}
\end{figure}

Considering the above observations, we conduct experiments to examine the influence of edge attributes with different time lengths. Fig.~\ref{fig:9} describes the effect of feature selection period with different durations in the edge autoencoder. Note that here we use only the edge autoencoder to avoid the influence of node attributes (i.e., academic age).
We observe that the overall trend of all metrics increases with more investigated years in the fields of Physics, Computer Science, and Economics. Particularly, the edge autoencoder can achieve better performance if we adopt the collaboration information of first 7 years. It is noticeable that the performance differs from one field to another. The reason accounting for this phenomenon is the differences in collaboration patterns between advisees and advisors which exist among the disciplines.

\subsection{Application and Visualization}
We apply Shifu2 onto the entire MAG dataset to generate academic genealogy automatically over all research fields. We know that some authors only publish one or two papers, and leave the academia at the early stage of their careers. In such cases, there is no long-term stable advisor-advisee relationship. How to select scholars with a stable advisor-advisee relationship is also a critical issue. Here we only apply Shifu2 to authors who meet the following criteria:
\begin{itemize}
  \item The author has published at least 1 paper every 5 years;
  \item The author has published at least 10 papers in the entire MAG dataset;
  \item The author's publication career spans at least 10 years.
\end{itemize}

We calculate the personal attributes and collaboration attributes as the input of the node autoencoder and the edge autoencoder, respectively. By applying Shifu2, we generate a large-scale advisor-advisee relationship dataset. It is an enrichment containing not only advisor-advisee pairs but also their academic attributes. This dataset can be used in many applications, such as supervisor finding, academic performance assessment, and reviewer recommendation.

\begin{figure}[htbp]
\centering
\subfigure[Chemistry]{
\label{fig:10-a}
\includegraphics[width=0.23\textwidth]{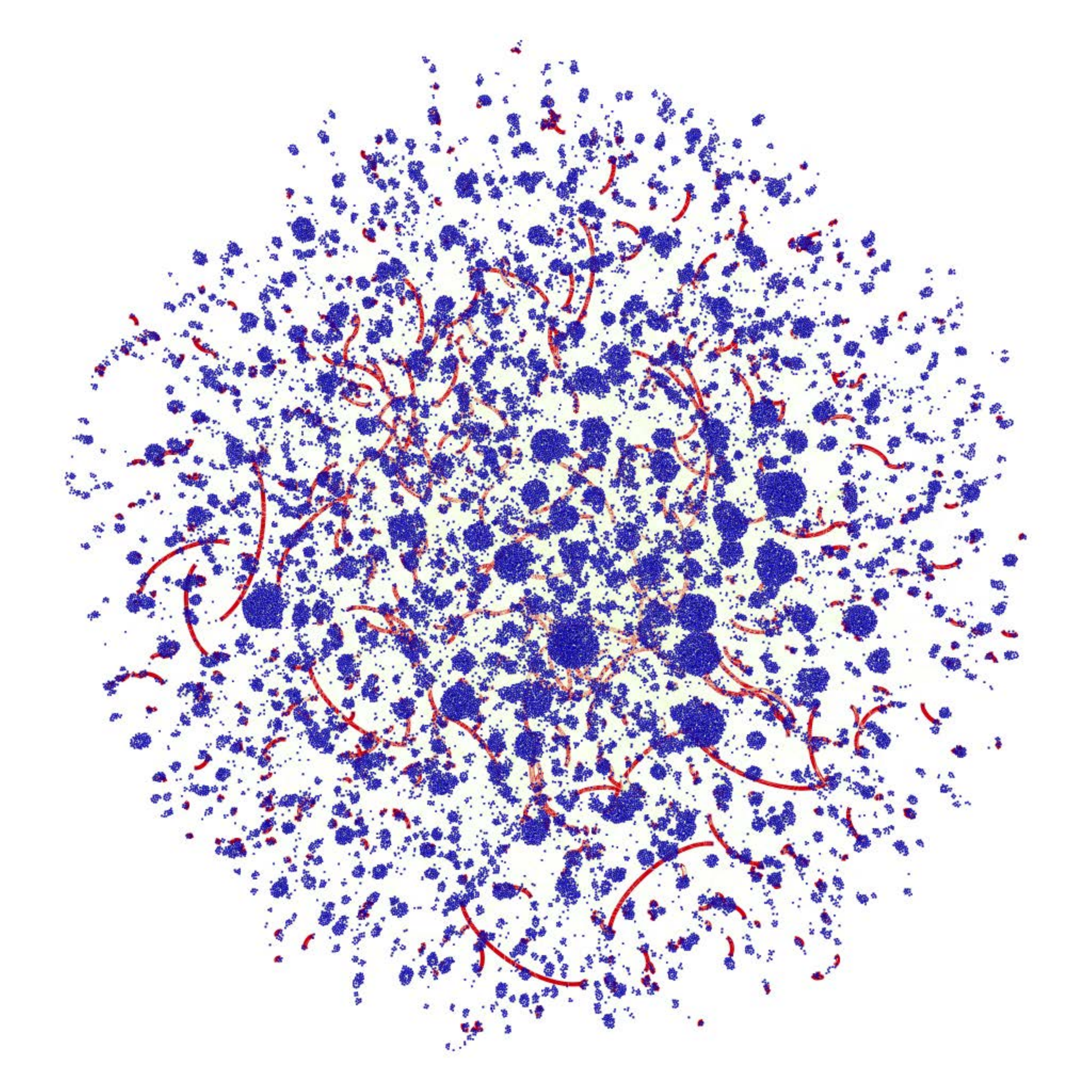}}
\subfigure[Computer Science]{
\label{fig:10-b}
\includegraphics[width=0.23\textwidth]{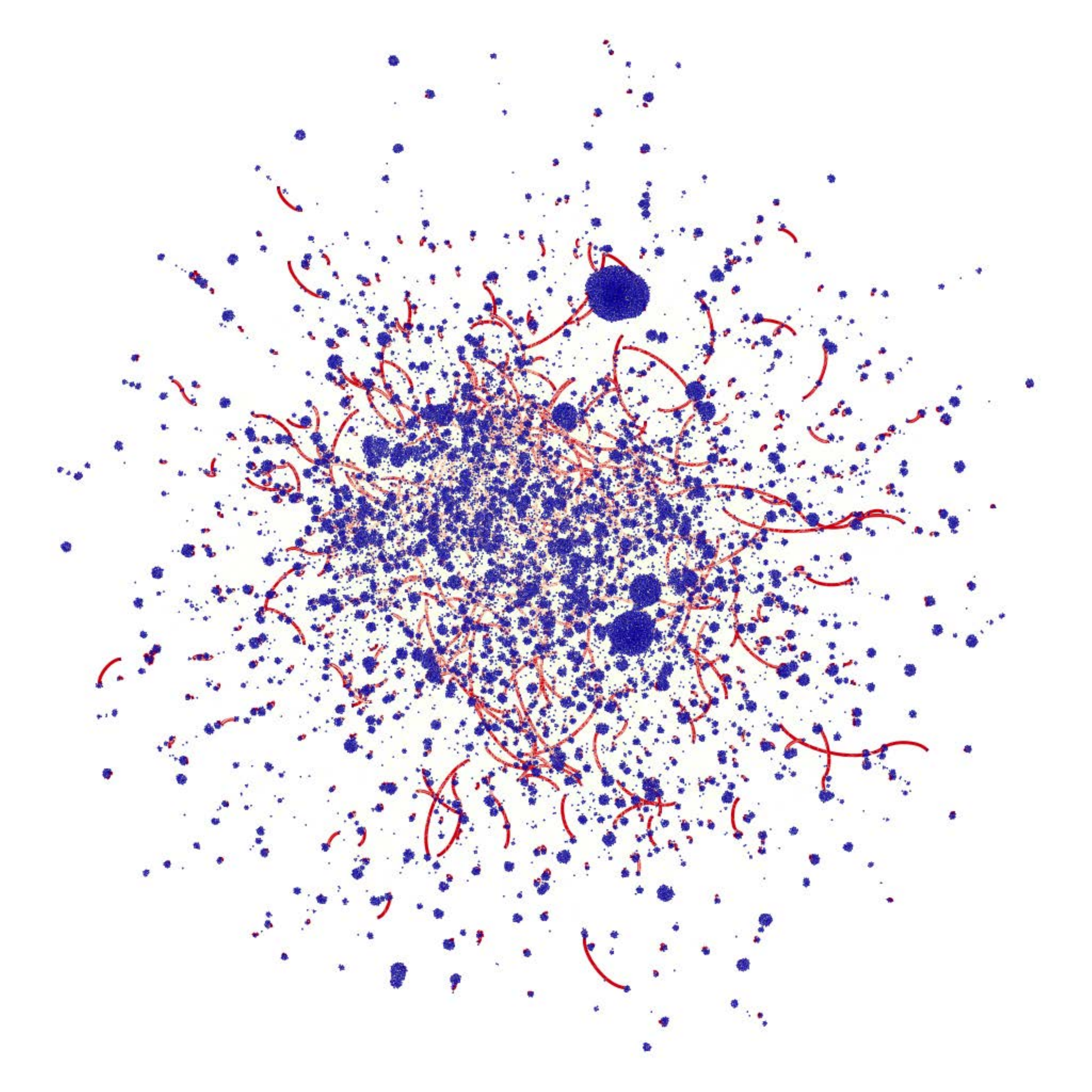}}\\
\caption{Collaboration networks in Chemistry and Computer Science.}
\label{fig:10}
\end{figure}
For example, Fig.~\ref{fig:10-a} and Fig.~\ref{fig:10-b} show the collaboration networks in Chemistry and Computer Science, respectively. Nodes represent advisees and their collaborators including their advisors. We use red edges to represent the collaboration between advisees and their advisors. We can see that the collaboration patterns between advisees and advisors are quite different in these two areas. The collaborations in Computer Science are more concentrated than Chemistry, which verifies the disciplinary differences in scholars' collaboration patterns.

\section{Conclusion}
\label{sec5}
In this work, we extract advisor-advisee relationships based on scholars' publication records. We have proposed an efficient NRL model called Shifu2 to model and identify the advisor-advisee relationship. Specifically, we transform the large-scale network into the low-dimensional space and learn the representations for both nodes and edges based on autoencoder. Experiments upon real scientific collaboration networks demonstrate the effectiveness and stability of the proposed model. Furthermore, we have applied Shifu2 onto the entire MAG dataset to generate the large-scale academic genealogy automatically.

The following topics could be good choices for future work in this line of research.

(1) Both MAG and AFT have limited accuracy and completeness. For future work, more practical problems, for instance, how to acquire high-quality advisor-advisee pairs for training, could be considered. A potential solution is using the whole ProQuest\footnote{https://search.proquest.com/index} dissertation dataset to extract ground-truth advisor-advisee pairs.

(2) How to extend Shifu2 for identifying other types of relationships such as friendship in social networks?

(3) The use of the obtained benchmark dataset in various applications is yet to be explored.

\section*{Acknowledgment}
This work is partially supported by National Natural Science Foundation of China (61872054) and the Fundamental Research Funds for the Central Universities (DUT19LAB23). Hanghang Tong is partially supported by NSF (IIS-1651203).

\bibliographystyle{IEEEtran}
\bibliography{IEEEabrv,AA}

\end{document}